\documentclass[10pt,twocolumn,letterpaper]{article}

\usepackage[accsupp]{axessibility}  

\usepackage{iccv}
\usepackage{times}
\usepackage{epsfig}
\usepackage{graphicx}
\usepackage{amsmath}
\usepackage{amssymb}

\usepackage{subcaption}
\usepackage{amssymb}
\usepackage{booktabs}
\usepackage{multirow}
\usepackage{arydshln}
\usepackage{xcolor}

\newcommand{\tablefont}{\small}
\newcommand{\boldheader}[1]{\noindent\textbf{#1.}}

\newcommand{\Name}{\textsc{CLOSE}}
\newcommand{\ImageModel}{\Name{} w/Images}
\newcommand{\Baseline}{\Name{} w/o Noise}
\newcommand{\NoiseModel}{\Name{}}
\newcommand{\TunedModel}{\Name{} w/Tuned Noise}

\newcommand{\coco}{\textsc{Coco}}
\newcommand{\cc}{\textsc{CC3M}}
\newcommand{\gptj}{\textsc{GPT-J}}
\newcommand{\vqat}{\textsc{VQA 2.0}}
\newcommand{\curie}{OpenAI Curie}
\newcommand{\vqae}{\textsc{VQA-E}}
\newcommand{\gptt}{\textsc{GPT-3}}
\newcommand{\cider}{CIDEr}
\newcommand{\meteor}{METEOR}
\newcommand{\spice}{SPICE}

\usepackage[pagebackref=true,breaklinks=true,letterpaper=true,colorlinks,bookmarks=false]{hyperref}

\iccvfinalcopy 

\def\iccvPaperID{11146} 

\ificcvfinal\fi

\begin{document}

\title{\emph{I can't believe there's no images! }\\Learning Visual Tasks Using Only Language Supervision}


\author{
Sophia Gu\thanks{Equal contribution} \quad \quad Christopher Clark\footnotemark[1] \quad \quad Aniruddha Kembhavi \\
Allen Institute for Artificial Intelligence \\
\tt\small \{sophiag, chrisc, anik\}@allenai.org
}
\maketitle


 \begin{abstract}

   Many high-level skills that are required for computer vision tasks, such as parsing questions, comparing and contrasting semantics, and writing descriptions, are also required in other domains such as natural language processing. In this paper, we ask whether it is possible to learn those skills from text data and then transfer them to vision tasks without ever training on visual training data. Key to our approach is exploiting the joint embedding space of contrastively trained vision and language encoders. In practice, there can be systematic differences between embedding spaces for different modalities in contrastive models, and we analyze how these differences affect our approach and study strategies to mitigate this concern. We produce models using only text training data on four representative tasks: image captioning, visual entailment, visual question answering and visual news captioning, and evaluate them on standard benchmarks using images. We find these models perform close to models trained on images, while surpassing prior work for captioning and visual entailment in this text-only setting by over 9 points, and outperforming all prior work on visual news by over 30 points. We also showcase a variety of stylistic image captioning models that are trained using no image data and no human-curated language data, but instead using readily-available text data from books, the web, or language models.
   

\end{abstract}

\section{Introduction}

Although vision and natural language processing (NLP) tasks are typically thought of as being very distinct, there is often a high degree of overlap in the skills needed to complete them.
Visual question answering and reading comprehension question answering both require parsing and understanding questions, visual entailment and textual entailment require comparing different semantic meanings, and captioning and summarization require writing text that summarizes the semantics of the input. 
This raises an intriguing possibility: if a model learned to complete one of these tasks using a high-level semantic representation of the input text, then in theory it could immediately be able to complete the corresponding visual task as long as the input image is encoded in the same semantic representation.
We call this challenge \textit{zero-shot cross-modal transfer} because it requires applying skills learned from one modality to a different one.
Achieving this would be a step towards building multi-modal models that can generalize skills across modalities without needing expensive training data for each modality, and has potential applications for tasks where visual training data is scarce but text data is relatively easy to collect.

Accomplishing this requires encoding images and text into a shared semantic space. We use vision and language (V\&L) models trained with a contrastive loss for this purpose~\cite{clip,align}. These models learn to embed text and images into vectors such that the vectors for matching images and captions are close together, and vectors for unrelated images and captions are far apart. Although this loss was originally intended for representation learning and zero-shot classification, here we show it also facilitates cross-modal transfer.

\begin{figure*}[th!]
    \centering 
    \includegraphics[scale=0.4]{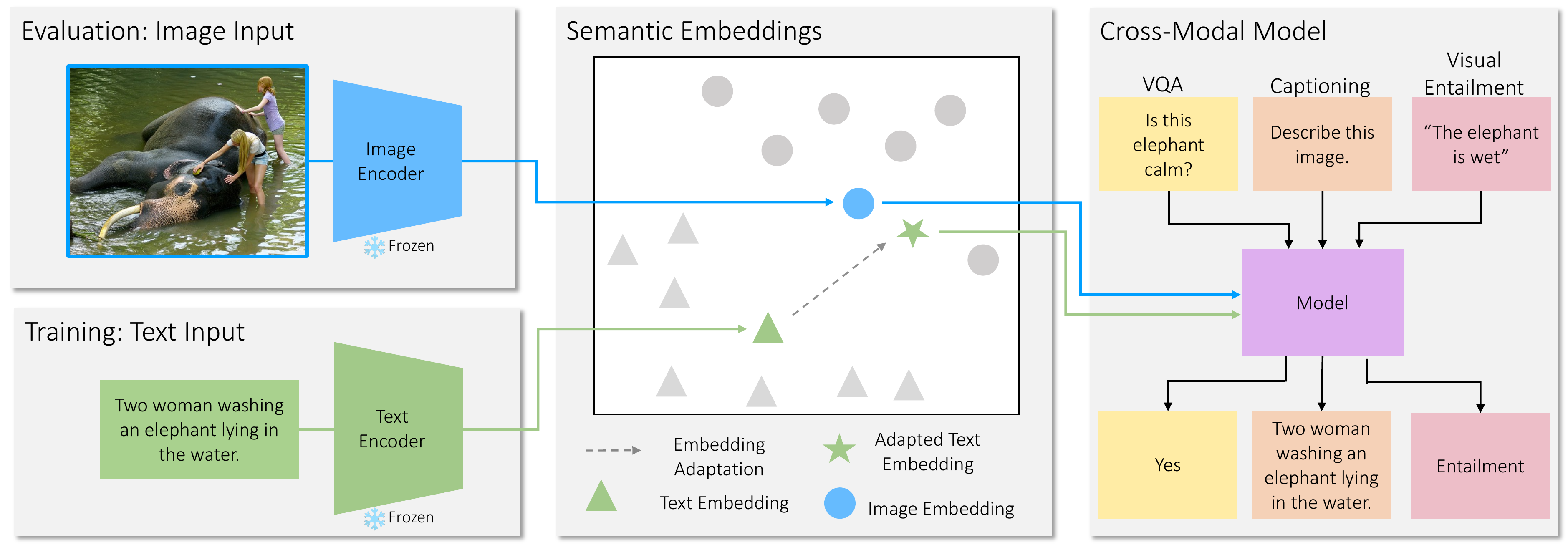}
    \caption{Overview of \Name{}. During training, input text is encoded into a vector with a text encoder and adapted with an adaptation method. A model learns to use the vector to perform a task such as VQA, captioning, or visual entailment.
    During testing, an input image is encoded with an image encoder instead to allow cross-modal transfer.}
    \vspace{-0.2cm}
    \label{fig:method}
\end{figure*}

To do this, we propose a method called Cross modaL transfer On Semantic Embeddings (\Name{}).
An outline of \Name{} is shown in Figure~\ref{fig:method}. During training, the text inputs are encoded into a vector using the (frozen) text encoder from a contrastive model, which is then used as an input to a model. During testing, the visual input is embedded with a (frozen) image encoder and used in place of the text embedding. Because these encoders were explicitly trained to produce embeddings that encode semantics in similar ways, learning to read and process the text vector should naturally translate to the ability to read and process the image vector. 
Although we focus on text-to-image transfer in this paper, our approach is applicable to other contrastive models such as videos~\cite{video_clip}, point clouds~\cite{afham2022crosspoint}, and audio~\cite{guzhov2022audioclip,elizalde2022clap,wu2022wav2clip}, potentially allowing transfer between many other modalities.

One potential difficulty with this approach is that, while contrastive embeddings do share some structure between modalities, there can still be significant differences between the image and text vectors in practice~\cite{Liang2022MindTG}. To mitigate this, we propose to additionally use \textit{adapters} that modify the text vectors being used during training. We find adding Gaussian noise to be very effective in boosting performance, but consider other approaches as well in our analyses.

Text-to-image transfer is a relatively unexplored setting, so we first conduct extensive experiments to establish that \Name{} can handle the text-to-image domain shift without a major performance drop.
We compare models trained with \Name{} on text alone to models trained with images and text on three standard V\&L tasks: captioning, visual questioning answers (VQA) and visual entailment, and the more complex task of visual news captioning~\cite{visual_news}. We find the text-only models generally perform reasonably close to versions trained with images, showing that \Name{} can effectively transfer many skills across modalities.
We surpass the previous best text-only method in captioning~\cite{esper} by 17 \cider{} (78.2 vs. 95.4) and visual entailment~\cite{song2022clip} by 9 points (66.6 vs. 75.9), making our method state-of-the-art for these settings by a large margin. 
There are no prior results for VQA and visual news in this setting, however we do surpass the previously best reported result in visual news even with images~\cite{visual_news} (50.5 vs 80.8 \cider{}).

\begin{figure}
    \centering 
    \includegraphics[width=0.5\textwidth]{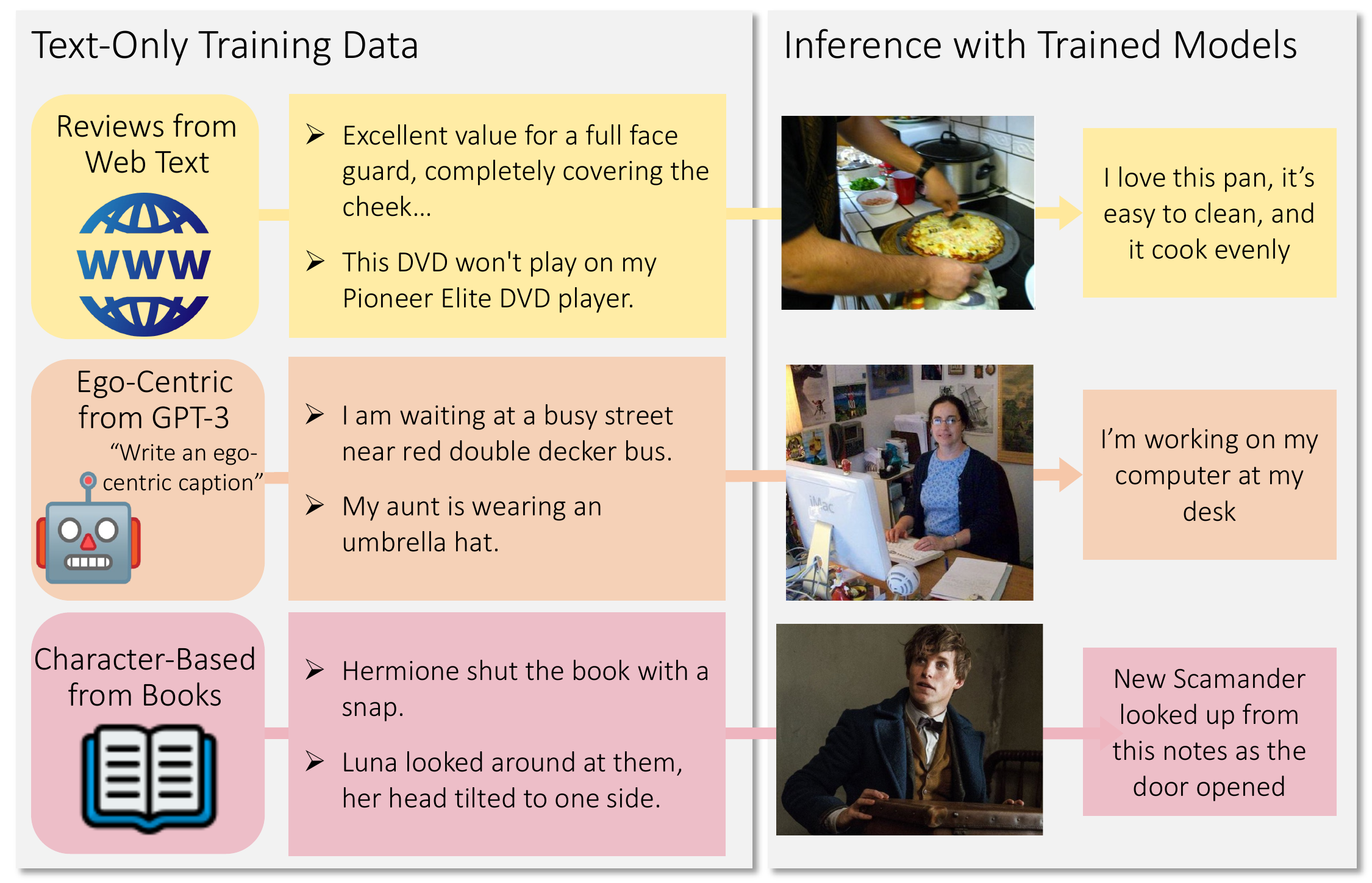}
    \caption{Using \Name{} to learn stylistic captioning without image data. Text examples of the desired style are gathered from sources such as the web, books, or \gptt{}. Models are trained on text only and then applied to images.}
    \label{fig:stylistic_captioning_teaser}
    \vspace{-0.5cm}
\end{figure}

These experiments show that efficient text-to-image transfer is possible. 
This has important practical implications because text training data can be directly constructed by annotators, mined from many existing text datasets, or even generated by a large language model such as \gptt{}~\cite{gpt3}, and can therefore be significantly less expensive than constructing visual training data.
We demonstrate this potential by training effective \Name{} captioning models from text generated by large language models~\cite{gpt3}, meaning the only human annotation required was for prompt construction.
We also train several stylistic captioning models without any labeled images (see Figure~\ref{fig:stylistic_captioning_teaser}).
We collect text with various styles from a diverse set of sources, including internet reviews, books, and \gptt{} generations, and demonstrate that \Name{} models trained on this text can produce accurate and stylistically correct captions for images.

Finally, we complete two analyses:
A sensitivity analysis showing that \Name{} is robust to cases where text and image vectors differ by a constant offset, which therefore allows \Name{} to work despite seemingly large differences between the image/text embeddings.
Additionally, a study on the effectiveness of using an auxiliary vision and language corpus to build an improved adapter. We find that improvements are possible but vary depending on the source of that data and that a particularly effective approach is to use the auxiliary data to compute a structured covariance matrix for use when adding Gaussian noise.


In summary, our contributions include: (i) introducing the \Name{} model for zero-shot cross-modal transfer; (ii) showing that training \Name{} with text data alone, on four V\&L tasks, gives results close to models trained on both images and text; (iii) SoTA results when using only text for three of the tasks; (iv) demonstrating an application of \Name{} for stylistic captioning; (v) analyzing how differences between image/text vectors in contrastive models and how different adapters affect \Name{}'s performance. To facilitate future work in the community, we release our code\footnote{\url{https://github.com/allenai/close}}.

%





\section{Method}
\begin{figure*}
    \centering
    \begin{subfigure}[b]{0.195\textwidth}
         \centering
         \includegraphics[width=\textwidth]{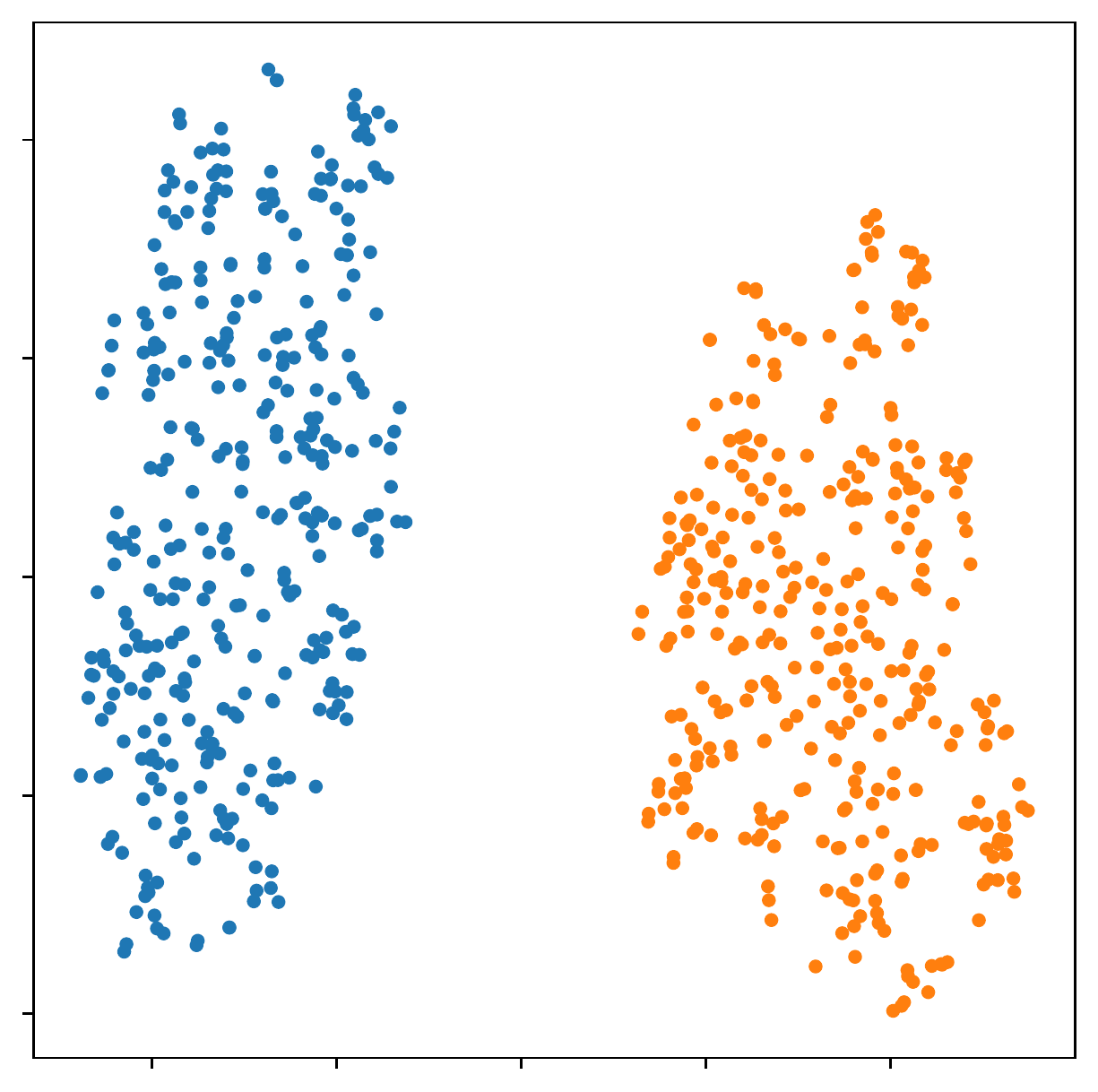}
         \caption{No Adapter}
         \label{fig:no_adapter}
     \end{subfigure}
    \begin{subfigure}[b]{0.195\textwidth}
         \centering
         \includegraphics[width=\textwidth]{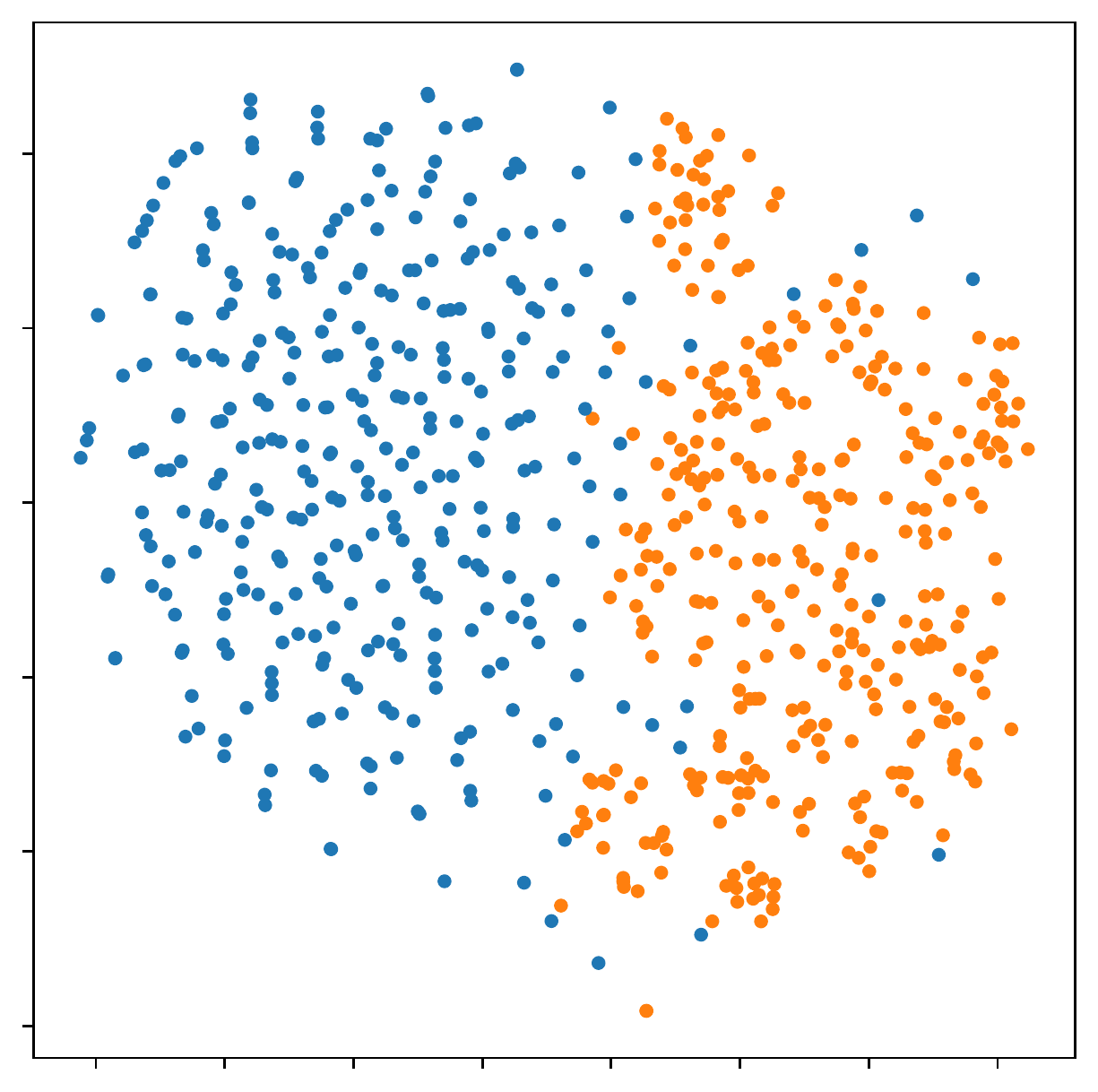}
         \caption{Gaussian Noise}
         \label{fig:noise}
     \end{subfigure}
    \begin{subfigure}[b]{0.195\textwidth}
         \centering
         \includegraphics[width=\textwidth]{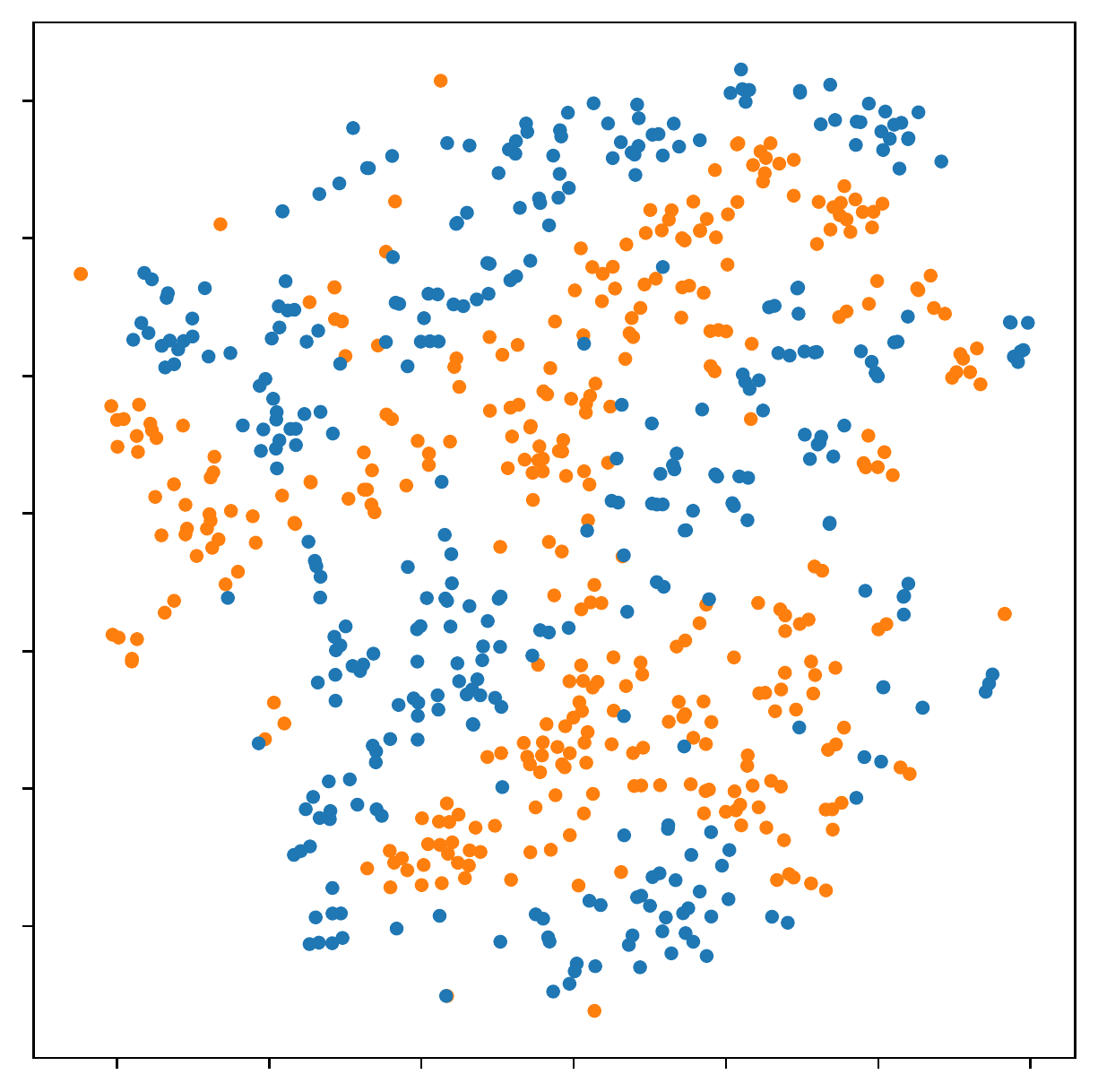}
         \caption{Mean Shift}
         \label{fig:mean_shift}
     \end{subfigure}
    \begin{subfigure}[b]{0.195\textwidth}
         \centering
         \includegraphics[width=\textwidth]{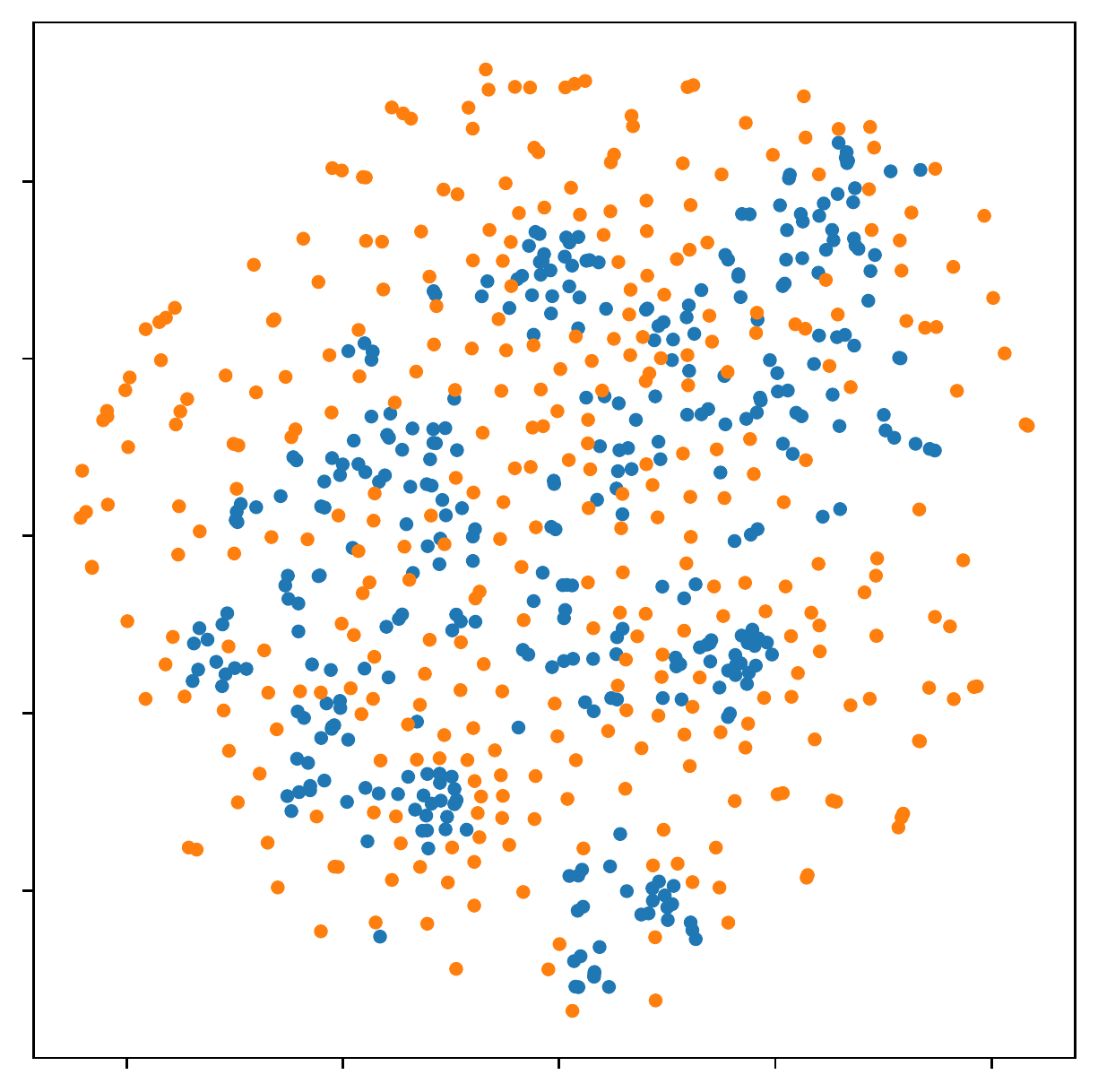}
         \caption{Mean Shift + Noise}
         \label{fig:shift_noise}
     \end{subfigure}
    \begin{subfigure}[b]{0.195\textwidth}
         \centering
         \includegraphics[width=\textwidth]{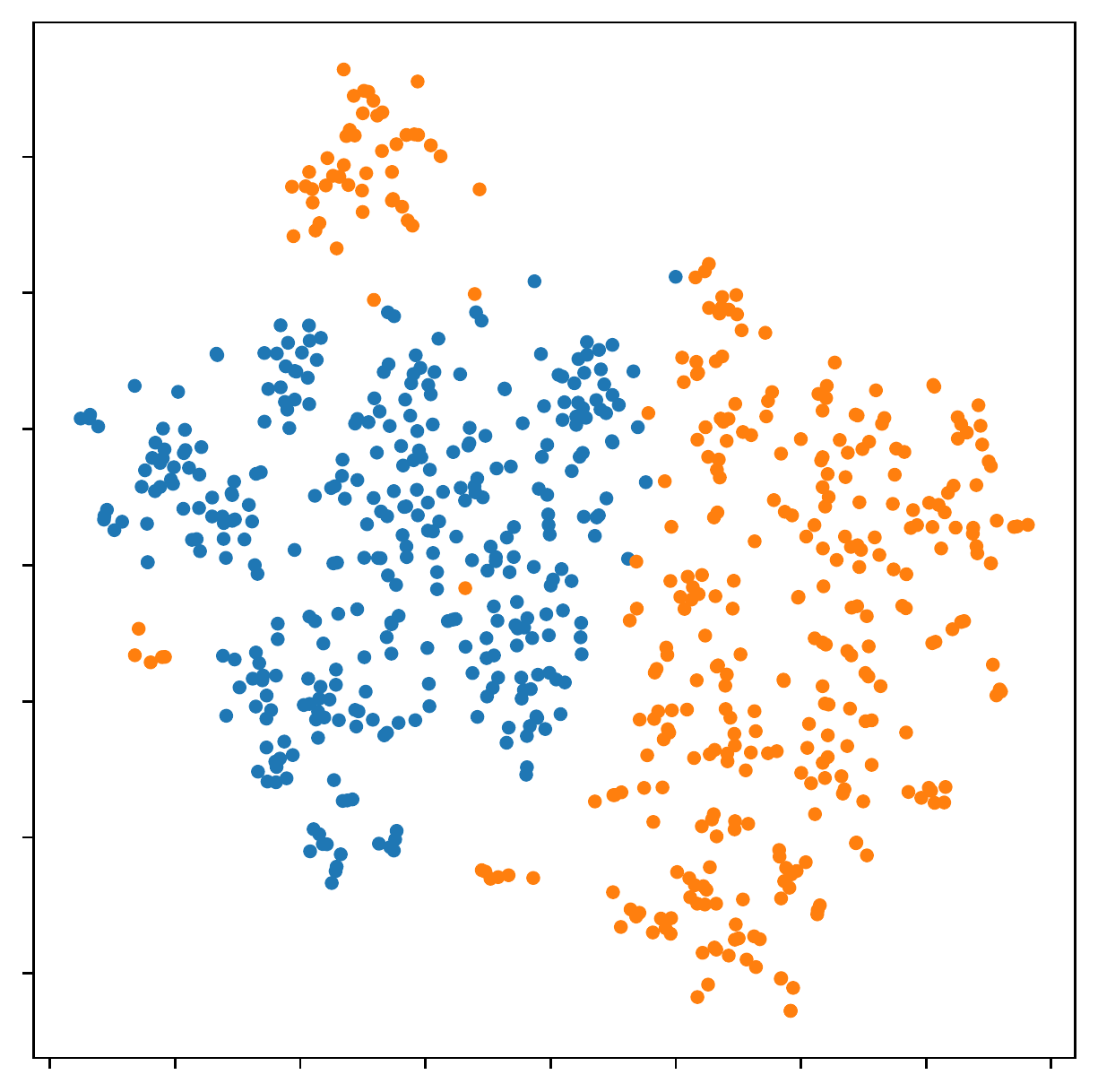}
         \caption{CC3M Mean Shift}
         \label{fig:cc3m_shift}
     \end{subfigure}
    \caption{t-SNE~\cite{van2008visualizing} plots for various adapters on 350 randomly selected image vectors (blue) and paired caption vectors (orange) from \coco{} captions. The first two panels demonstrate \Name{}, and the remaining three show additional adapters we study in our analysis (Section ~\ref{sect:analysis}).}
    \label{fig:modality_gap_plots}
\end{figure*}

\noindent \textbf{Model.}  
Our approach uses the image/text encoder from a contrastive model to encode the input, and then follows many prior works (e.g.,~\cite{Kamath2022WeblySC,Chen2019UNITERLU}) by fine-tuning a pre-trained language model to process the input vector, along with any additional input text, to generate output text.
First, the input image or text vector is normalized to have unit length to match what is used in the contrastive loss.
Then that vector is converted into a number of vectors, we use 4 in our experiments, of the same dimensionality as the language model's embedding layer using a linear layer.
Next, other input text (e.g., the hypothesis in visual entailment or the question in VQA) is tokenized and embedded with the language model's embedding layer. Those embeddings are concatenated with the embeddings built from the input vector to construct an input sequence for the language model.

For the sake of simplicity, we train the model generatively for all tasks~\cite{Gupta2021TowardsGP,Cho2021UnifyingVT}. The model generates a caption, a free-form question answer, or a class name for the tasks of captioning, VQA, and visual entailment respectively. During training, the language model and linear layer are fine-tuned, but the text encoder is kept frozen to ensure the correspondence between text and image vectors learned during pre-training is preserved.

\label{sect:modality_gap}
\noindent \textbf{Modality Gap.}
In practice, text and image vectors from contrastive models can be far apart, a phenomenon known as the modality gap~\cite{Liang2022MindTG}. For example, on \coco{} captions~\cite{coco_captioning} the average cosine similarity between an image and paired caption is only 0.26, while the average similarity between two unrelated captions is 0.35. Figure~\ref{fig:no_adapter} shows this gap causes image and text vectors to fall into separate clusters in the vector space.
The root cause is that the cross-entropy loss used by contrastive models only requires paired image and text vectors to be close \textit{relative} to random image and text pairs, which does not necessarily mean they are close in absolute terms, see Liang \etal~\cite{Liang2022MindTG} for more discussion.




We thus adopt a simple and effective solution -- adding Gaussian noise that is drawn from a standard normal distribution and then scaled by a hyper-parameter $w$, to the text vectors during training. Intuitively, this noise helps to close the modality gap by spreading out the text vectors and overlapping them with the image vectors. 
Figure~\ref{fig:noise} visually shows that even a small amount of noise leads to much better overlapping of the image and text vector spaces.
The noise also encourages the model to be more robust to minor changes or variations to the input vectors, and thus be better prepared for the shift caused by switching from text to image vectors. 

A second motivation for using random noise is the observation that image vectors capture certain subtle visual details like lighting, background, or camera position that are not reflected in the text vectors. To illustrate this, we show a small case study in Appendix 5 where we observe that semantic changes (e.g., changing the subject of a caption or image from ``dog" to ``cat") result in a relatively consistent directional shift for text vectors, but has a more erratic effect on image vectors. 
Adding noise to the text embedding helps to mitigate this problem by simulating the fact that, even for semantically similar inputs, image and text vectors can still have minor differences due to the additional information encoded in the images.

After adding the noise we re-normalize the vector to unit length to match the image vectors that will be used during evaluation. 
We study the modality gap and other approaches to handling it in more detail in Section~\ref{sect:analysis}.





\section{Experiments}
\begin{table*}[]
    \tablefont
    \centering
    \begin{tabular}{l c c c c c c c}
        \toprule
        Model & Text-Only & Cap. (Single) & Cap. (Mult.) & VE & VQA & E-VQA & VN \\
        \midrule
        \multirow{2}{*}{Prior Work} & \multirow{2}{*}{\checkmark} & \multirow{2}{*}{-} & ESPER Style~\cite{esper} & CLIP Cls.~\cite{song2022clip}  &  TAP-C~\cite{song2022clip} & \multirow{2}{*}{-} & \multirow{2}{*}{-}  \\
         & & & 78.2 & 66.6 & 38.7 & &  \\
        \midrule
        \Name{} w/o Noise & \checkmark & 16.4 & 68.7 & 68.2 & 60.2 & 59.8 & 32.1 \\ 
        \textbf{\Name{} (Ours)} & \checkmark & 80.5 & 95.3 & 75.9 & 59.6 & 62.9 & 80.8 \\
        \midrule 
        \Name{} w/Tuned Noise & & 95.4 & 98.4 & 75.9 & 61.9 & 64.3 & 80.8 \\
        \Name{} w/Images & & 113.2 & 113.2 & 77.7 & 65.4 & 67.9 & 105.7\\ 
        \bottomrule
    \end{tabular}
    \caption{Results on V\&L tasks. Models in the last two rows require images and so are upper bounds for \Name{}. We report \cider{}~\cite{cider} for captioning with single and multiple captions, visual entailment test accuracy, VQA 2.0 test-dev accuracy, E-VQA validation accuracy, visual news test \cider{}.
    See Appendix 2 for other metrics and more detailed results. 
    }
    \label{tab:joint}
\end{table*}

We report results on four V\&L tasks: captioning, visual entailment, VQA and visual news, and when training \Name{} using only text generated by a language model. 

\subsection{Setup}
We construct pure-text training datasets for these tasks using the text annotations from the relevant training datasets, and, for some tasks, text captions of the training images.
Our primary point of comparison is a \Name{} model trained with the training images, in which case the images are encoded with the image encoder during training in the same manner as done during testing. This model does not experience domain shift, so we view it as an \textit{upper bound}.
We emphasize that in practice the text training data could come from many other possible sources, see Sect.~\ref{sect:stylistic_captioning} and Sect.~\ref{sect:training_with_language_model_data} for additional experiments that demonstrate this, we use these text sources since they closely match the data the models with images are trained on and therefore allow us to better isolate and study what performance is lost due to the image-text domain shift.

We use T5$_{base}$~\cite{t5} and CLIP$_{ViT-L/14}$~\cite{clip}, a noise level of 0.08, and a fixed set of hyper-parameters for all tasks to demonstrate our method is effective even when there is no image/text validation set to tune on. See Appendix 1 for hyper-parameter details.  
We additionally show results when the noise level is tuned on validation sets, and when the noise is removed, to study the effect of noise on \Name{}.




\subsection{Results}
Results are shown in Table~\ref{tab:joint}. Due to space constraints, we only report one metric for each task here and include more results in Appendix 2. 
We also show the best method from prior work, when present, that does not use images. 

\boldheader{Image Captioning}
\label{sect:captioning_experiments}
For captioning, we use text captions as both the input text and the target output text.
However we find that, if multiple captions about one scene are available, it is beneficial to use different captions about the same image as the input and target text. We call the first setting \textit{captioning (single)} and the second \textit{captioning (multiple)} and evaluate both since they facilitate different training setups.
We evaluate on \coco{} Captioning~\cite{coco_captioning} using the Karpathy split~\cite{karpathy2015deep}. We train our text-only models using just the captions in the training data. We treat all captions per image as a group for the multiple-caption setting and use each caption individually in the single-caption setting.


\Name{} reaches 95.3 \cider{} in the multiple caption setting, showing high captioning competency despite not using images. In the single caption setting, performance is reduced but can be increased to 95.4 with higher noise levels.
Our approach is substantially better than recent zero-shot methods such as MAGIC (49.3)~\cite{tewel2021zero} and Socratic Models (44.5)~\cite{zeng2022socraticmodels}, and is 17 points ahead of ESPER Style (78.2)~\cite{esper} which also uses text captions.

\boldheader{Visual Entailment}
Visual entailment requires determining whether a premise image either entails, contradicts, or is neutral with respect to a hypothesis sentence.
During training, a text premise is used instead of an image. The hypothesis sentence is always text and is encoded with the language model. We train on SNLI~\cite{snli} (a language-only dataset) and evaluate on SNLI-VE~\cite{visual_entailment} (a vision and language dataset).
Despite not using images, \Name{} achieves similar performance to the image model. Song \etal~\cite{song2022clip} also experiment with this task, but we find adding Gaussian noise allows us to surpass their result by over 9 points.

\boldheader{VQA}
To train a VQA model we use data that contains a sentence describing a scene (encoded with the text encoder), a question (encoded with the language model), and a target answer. We consider two datasets. First, we pair \coco{} captions with questions about the same image from \vqat{}~\cite{vqa2}. However, in this dataset, the questions might ask about details of the image not included in the caption, and thus cannot be answered by the text-only model. Hence we also train and evaluate on \vqae{}~\cite{vqa_e} which contains a subset of the \vqat{} questions paired with \coco{} captions that have been verified to contain the answer. 

These training sets have significantly different question distributions due to the filtering done in \vqae{}, so we evaluate models either on the \vqat{} test-dev set or the \vqae{} validation set\footnote{\vqae{} does not have a test set} depending on what train set was used.
There is no prior work for this task in the text-only setting, however \Name{} does outperform TAP-C$_{ViT-B/16}$~\cite{song2022clip}, a CLIP-based zero-shot approach.

For \vqae{}, we observe only a 3.5 point drop in accuracy relative to image training while surpassing the baselines. The gap is more significant on \vqat{}, which we attribute to the sometimes poor alignment between the captions and questions, although our method is still within 5 points of the model trained on images.

\boldheader{Visual News}
Visual news requires captioning an image in the context of a news article, and which therefore often requires mentioning the people, locations, and events from the article text~\cite{visual_news}. \Name{} is easily extended to this setting by using the caption as both the image text and the target output, while the article is given as additional context to the language model. For this task, we randomly sample 15\% of the training data each epoch due to the large dataset size, and use OpenCLIP instead of CLIP since our previous experiments found it slightly improves performance.
\Name{} with images achieves over 105 \cider{}, a significant improvement over the previous best benchmark of 50.5 \cider{}~\cite{visual_news}. Training without images also outperforms the previous state-of-the-art, obtaining a respectable 80.8 \cider{}. See Appendix 5 for qualitative examples.

\boldheader{Discussion}
Overall, performance is comparable to the model trained with images showing \Name{} is able to transfer skills between modalities. Tuning the noise level can benefit some tasks, therefore better heuristics for choosing the noise level or leveraging a small image/text validation set could additionally improve performance. On the other hand, removing the noise reduces performance drastically across almost all tasks. This is because the noise plays an important role in addressing the modality gap.


\subsection{Training with Data from a Language Model}
\label{sect:training_with_language_model_data}
\begin{figure}
    \centering
    \includegraphics[width=0.45\textwidth]{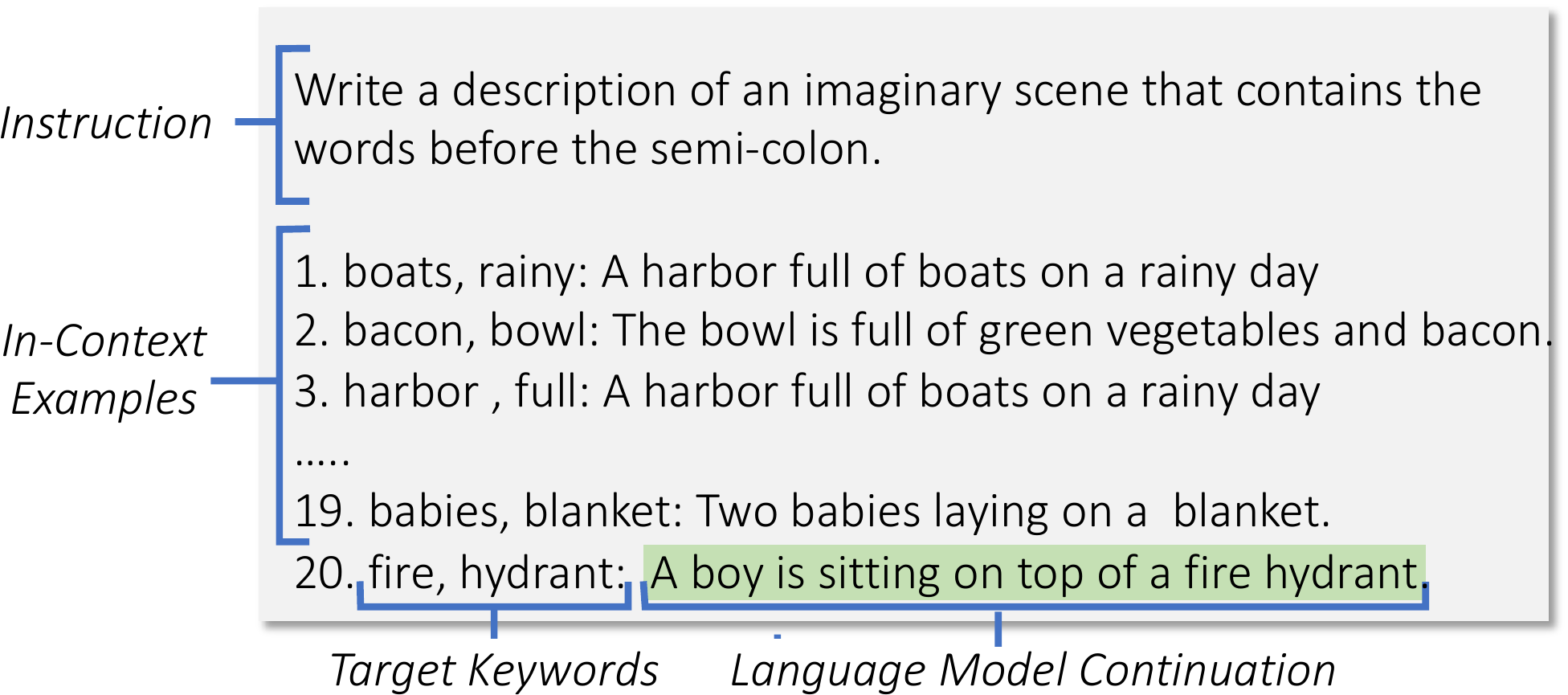}
    \caption{Prompt used to generate a synthetic caption from a language model. The language model's continuation (highlighted text) is used as a synthetic caption.}
    \vspace{-0.2cm}
    \label{fig:synthetic_caption_figure}
\end{figure}

\begin{table}[]
    \tablefont
    \centering
    \begin{tabular}{l c c c c}
        \toprule
         Model & B-4 & M & C  & S \\
        \midrule
        MAGIC~\cite{su2022language} & 12.9 & 17.4 & 49.3 & 11.3 \\
        \Name{} w/\coco{} & 29.5 & 25.6 & 98.4 & 18.3 \\ \hdashline
        \Name{} w/\gptj{} RNG & 19.6 & 20.9 & 63.2 & 13.8 \\
        \Name{} w/\gptj{} Unigram & \textbf{23.2} & \textbf{22.2} & \textbf{78.9} & \textbf{15.6} \\
        \Name{} w/OpenAI Curie & 18.5 & 21.2 & 69.0 & 14.9 \\
    \bottomrule
    \end{tabular}
    \caption{BLEU-4, \meteor{}, \cider{}, and \spice{} on the \coco{} validation set when training on synthetic captions.}
    \label{tab:synthetic_captioning}
\end{table}


Next, we use \Name{} to train a captioning model on synthetic data generated by a language model. We first construct a prompt that includes a natural language instruction and some example captions following an in-context learning approach~\cite{gpt3}, shown in Figure~\ref{fig:synthetic_caption_figure}.
To generate a diverse set of captions, we prefix each caption with two keywords that occur in that caption, and end the prompt with two new keywords to be used in the caption to be generated (``fire" and ``hydrant" in Figure~\ref{fig:synthetic_caption_figure}). 
Then diverse captions can be constructed by changing the ending keyword pair. To reduce the chance of caption style affecting the quantitative evaluation, we take steps to better match the style of the \coco{} captions, although in settings where the precise style is of less importance this would not be required.
We generate 100k examples from three generation methods:

\boldheader{\gptj{} RNG} Examples are generated using a 6 billion parameter open source language model, GPT-J\cite{mesh-transformer-jax}, with 50 in-context examples. Keywords are sampled uniformly at random from keywords in the \coco{} training data.

\boldheader{\gptj{} Unigram} Keywords are instead sampled to match the unigram distribution of \coco{} captions.

\boldheader{Curie Unigram} Generations are from OpenAI Curie\footnote{\url{https://beta.openai.com/docs/models/gpt-3}} with 20 examples and unigram-matching.

Results on \coco{} are shown in Table~\ref{tab:synthetic_captioning}. 
Our best result achieves 78.9 \cider{}. Inspection shows that, even with our keyword sampling approach, many errors are still caused by style issues, and that style also explains the reduced performance of the Curie model. For example, the synthetic captions from the Curie model are 23 times more likely than the \coco{} and the GPT-J captions to use the word ``opens" (e.g., ``a living room that opens onto the balcony"), and use ``cellphone" while ``cell phone" is much more common in \coco{}. 
More details are in Appendix 3. 
This illustrates how, when using this method, the choice of language model can have subtle effects on the style of captioning that will be learned. 
Despite this issue, this is still a very strong result that surpasses the zero-shot method MAGIC~\cite{su2022language}.

\section{Analysis}
\label{sect:analysis}
Our approach opens up two intriguing questions: (1) Why does embedding substitution work even when text and image vectors are generally quite far apart? (2) Can methods that leverage additional data to better close the modality gap improve upon this approach? We do two analyses to answer these questions. 
Furthermore, we study how different choices for the contrastive embedding model or for the language model affect our method's performance.

\subsection{Sensitivity Analysis}
\label{sect:sensitivity_analysis}
\begin{table}[]
    \tablefont
    \centering
    \begin{tabular}{l | c c c | c c c} \toprule
        Bias & Mag. & MG & $\Delta$ & Cap. & VE & VQA  \\
        \midrule
none & 0.0 & 0.26 & 1.00 & 94.4 & 64.3 & 75.9 \\ \hdashline
mean & 0.8 & 0.62 & 0.69 & 92.8 & 64.7 & 75.4 \\
-mean & 0.8 & -0.10 & 0.85 & 84.3 & 62.0 & 71.8 \\
RNG & 0.2 & 0.25 & 0.98 & 93.5 & 63.9 & 75.3 \\
RNG & 0.5 & 0.24 & 0.89 & 92.5 & 64.2 & 75.3 \\
RNG & 0.8 & 0.20 & 0.78 & 89.3 & 63.7 & 74.8 \\
RNG & 1.0 & 0.18 & 0.71 & 87.2 & 63.8 & 74.2 \\
RNG & 2.0 & 0.11 & 0.45 & 73.7 & 61.4 & 71.3 \\

\bottomrule
    \end{tabular}
    \caption{Text vector translation-sensitivity analysis. The first three columns show the translation magnitude, the resulting modality gap on \coco{}, and the cosine similarity to the original vectors.
    The following columns show \cider{} captioning score, accuracy on \vqae{}, and accuracy on visual entailment on validation sets.}
    \vspace{-0.3cm}
    \label{tab:sensitivity_analysis}
\end{table}
To help answer the first question, we perform a sensitivity analysis on the input text vectors.
To do this, the model is trained while adding a constant vector to the normalized text vectors and then re-normalizing, and tested on the unaltered image vectors as before. 
This alteration will change how the text vectors are distributed relative to the image vectors, but will not change how the text vectors are distributed relative to one another.
We show results when using a random vector (note the same vector is used for all of training, it will just be selected randomly at the start of training) of different magnitudes, the mean difference of text and image vectors to represent a shift towards the image vectors, and the negation of that vector to shift away from the image vectors. In all cases, we continue to add Gaussian noise as before.

Results are shown in Table~\ref{tab:sensitivity_analysis}. For random vectors (RNG), we report the average of three runs with 3 different vectors.
Overall, we see only minor degradation when using random vectors until very large shifts are used, showing the model is generally insensitive to shifting the text vectors during training. 
Shifting the vectors towards the images (mean) can result in a slight gain in performance, and shifting the vectors away from them (-mean) results in a more significant decrease, showing the model is not completely insensitive. However it is still notable that vector substitutions work well even as the text vector's positions are significantly randomized.

We hypothesize that this insensitivity is due to two reasons. First, most directions in the shifted feature space are predictive of the output in the same manner as before because the text vectors do not change relative positions. Second, the Gaussian noise trains the model to be insensitive to shifts in unimportant directions in the feature space, which often include the direction of the shift. 
This insensitivity provides part of the answer to question 1. A major source of the modality gap is a constant shift between the image and text vectors~\cite{Liang2021CrossModalGL}. However, addressing this is not as important as one might expect because \Name{} is not highly sensitive to the absolute positioning of the text vectors.

\begin{figure*}[h]
    \centering
    \includegraphics[width=1.0\textwidth]{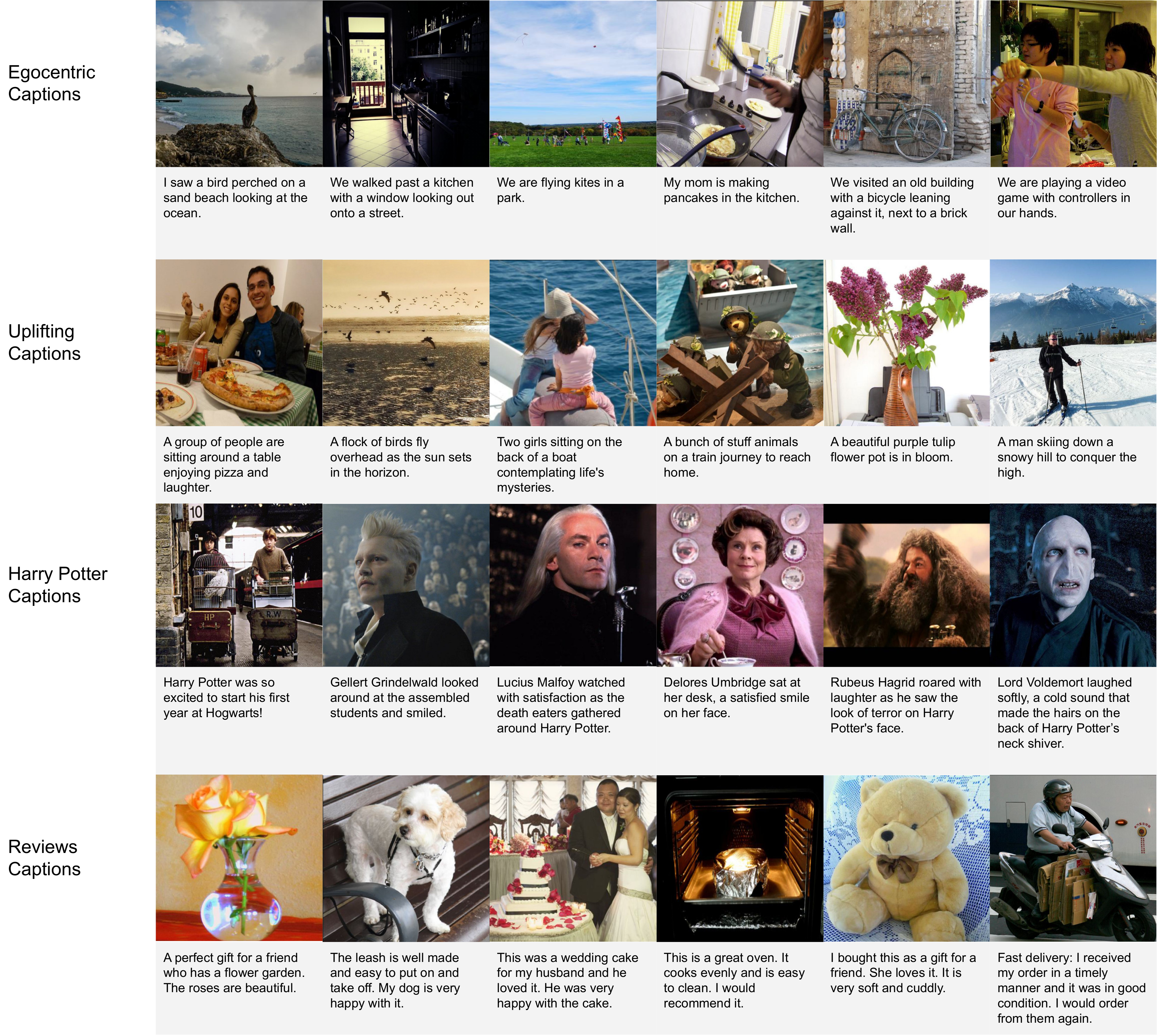}
    \caption{Examples of stylistic captions produced by \Name{} trained with only text data, and then applied 0-shot to images.}
    \vspace{-0.2cm}
    \label{fig:stylistic_captioning}
\end{figure*}

\subsection{Learned Adapter Analysis}
\label{sect:ablation}

\begin{table}[]
    \tablefont
    \centering
    \begin{tabular}{l | c | c c c c}
        \toprule
        Method & MG & Cap. & VE & VQA & VN \\ \midrule
        \NoiseModel & 0.26 & 94.3 & 75.9 & 64.3 & 80.8 \\
        +Cov. (\coco{}) & 0.62 & \textbf{106.5} & 75.5 & 65.5 & \textbf{84.1} \\
        +Cov. (\cc{}) & 0.58 & 95.1 & 75.8 & 65.0 & - \\
        +Linear  (\coco{}) & 0.81 & 99.5 & \textbf{76.0} & \textbf{65.7} & - \\
        +Linear (\cc{}) & 0.75 & 81.8 & 75.5 & 64.9 & - \\
    \bottomrule
    \end{tabular}
    \caption{Results with adapters built with paired data. The modality gap on \coco{} captions, captioning \cider{}, visual entailment accuracy, \vqae{} accuracy and visual news CIDEr are shown. The last task is more complex and so we only experiment it with one promising adapter.}
        \label{tab:ablation} 
\end{table}
As suggested by Figure~\ref{fig:mean_shift}, mean shift might not be perfect at aligning the text and image vectors, so we hypothesize more sophisticated adaption methods could improve performance. 
More complex adapters generally require a paired image/text corpus to train on, so we avoid using them in our main \Name{} method. However, here we investigate them to better understand how much performance they could potentially contribute.  
To study the difference between using high-quality annotated data or web data we use both \coco{} captions and Conceptual Captions 3 Million (\cc{})~\cite{cc3m}. For \coco{} we use the 30k captions from the ``restval" set of the Karapathy split, which do not appear in our train, eval or test sets, and for \cc{} we use a random sample of 100k image/text pairs. We consider two adapters:

\boldheader{Linear Adapter}
We learn the modality shift by training a linear model to minimize the Euclidean distance between the adapted text vector and its paired image vector. We continue to add Gaussian noise after applying this model.

\boldheader{Structured Noise with Covariance Matrix}
\label{sect:cov}
Even in principle, we do not expect there to be a perfect one-to-one mapping between text and image vectors because an image vector can be similar to many different texts that describe different parts or details of the image. This motivates us to approach the problem from the perspective of better understanding how text vectors are \textit{distributed} around its related image vectors, instead of just trying to learn a simple mapping function. 
In Appendix 4, we provide insight into how the vector differences from \coco{} image-caption pairs follow a particular shape.
To capture this shaped relationship between text and images, we add Gaussian noise whose mean and covariance are learned from the differences between text-image vectors in the auxiliary corpus, to the text during training.
This noise is expected to better simulate the text-image shift that will occur during evaluation.




Results are shown in Table~\ref{tab:ablation}. We observe large improvements on captioning, modest improvements on VQA and visual news\footnote{We only test one adapter on this task due to the longer training times}, and similar performance on visual entailment using the adapters from \coco{}, with the structured noise approach being significantly better on captioning, and slightly worse on the other tasks. 
The \cc{} adapter also achieves mild gains, although it is less effective. This shows the training data used for the adapter is important, a point that can be qualitatively observed in Figure~\ref{fig:mean_shift} and Figure~\ref{fig:cc3m_shift}.

\subsection{Performance Analysis of Different CLIP and T5 Models}
\label{sect:ablation-results}

\begin{table}[h!]
    \tablefont
    \centering
    \begin{tabular}{l c c c c c}
        \toprule
         CLIP Model & T5 Model & Cap. & VE & VQA\\
        \midrule
        ViT-L/14 & small & 94.4 & 74.9 & 59.9 \\
        ViT-L/14 & base & 95.4 & 76.1 & 64.3 \\
        ViT-L/14 & large & 93.9 & 75.1 & 65.2\\
        ViT-B/32 & base & 91.1 & 75.3 & 61.4 \\
        RN101 & base & 90.0 & 75.4 & 59.8\\
        RN50 & base & 90.2 & 75.3 & 60.4\\
        RN50$\times{4}$ & base & 92.0 & 75.3 & 61.5 \\
        RN50$\times{16}$ & base & 93.4 & 74.4 & 62.5\\
        RN50$\times{64}$ & base & 96.1 & 75.8 & 64.2\\
        OpenCLIP~\cite{openclip} & base & 99.2 & \textbf{76.3} & 65.1\\
        EVA-CLIP~\cite{eva} & base & \textbf{101.7} & 75.53 & \textbf{66.6}\\
    \bottomrule
    \end{tabular}
    \caption{Ablations with different contrastive and language models. The first column indicates which CLIP model was used, with OpenCLIP indicating we use the ViT-L/14 OpenCLIP model trained on Laion 400m~\cite{openclip}. The last three columns show \cider{} on \coco{} captioning in the single caption setting, accuracy on visual entailment, and overall accuracy on \vqae{} on the validation sets.}
    \label{tab:supp_captioning}
\end{table}

Finally, we study how different choices for the contrastive embedding model or for the language model affect the performance of our method.
Results for captioning, visual entailment, and E-VQA are shown in Table~\ref{tab:supp_captioning}. For these experiments we use the tuned noise values in order to compare best-case performance. We find the optimal noise level for these models generally does not change as these components are altered, so we use the same noise levels as our main results for all these experiments.

There is a consistent decrease in performance when using CLIP versions other than ViT-L/14, with only RN50×64 being comparable, showing that \Name{} gains effectiveness as the contrastive model becomes more powerful. We also observe much less dependence on the size of the T5 model, with the large model increasing performance on VQA but not on the other tasks. The OpenCLIP model is generally more effective and boosts the captioning results to nearly 100 \cider{}. The EVA-CLIP model~\cite{eva} further boosts VQA scores, approaching our main result with images (67.9), showing that \Name{}'s performance can be improved by enhancing the contrastive model.


\section{Stylistic Captioning}
\label{sect:stylistic_captioning}
We demonstrate an application of our method by applying it to the task of constructing captions with specific writing styles. Our general approach is to gather text-only training data that exemplifies the style we want the model to use, train on them as if they were text captions as done in Section~\ref{sect:captioning_experiments}, and then apply the model to images.
To show that a diverse range of natural language data sources can be used to learn different styles we show four captioning styles, each of which uses a different method of collecting training data.

\boldheader{Ego-Centric} 
Section~\ref{sect:training_with_language_model_data} shows that our model can be trained using data generated by a language model. 
Now we demonstrate an application of that approach by using the language model to generate captions in an ego-centric style. We use the same prompt format as before (Figure~\ref{fig:synthetic_caption_figure}), only now with 20 examples of manually authored captions written from a first-person perspective. We again sample keywords randomly from those found in \coco{} training captions to generate diverse prompts and obtain 20k captions using OpenAI's GPT-3 model. We apply this model to \coco{} validation images, shown in the top row of Figure~\ref{fig:stylistic_captioning}, and observe it learns to use a variety of first-person language while accurately describing the image.

\boldheader{Uplifting}
We use a publicly available dataset~\cite{Gan2017StyleNetGA} to collect 6k examples of uplifting captions (no images). Results are shown in the second row in Figure~\ref{fig:stylistic_captioning}, where we observe the model adds warm and optimistic details to its captions.
\\
\boldheader{Character-Based}
Next, we target character-based captions that use proper nouns and describe images as if they were from a story. Using proper nouns would be a significant hurdle for many existing systems due to the lack of image/name paired data in existing datasets.
However, \Name{} can leverage CLIP's ability of recognizing names of famous people~\cite{clip} to handle that problem.
We first pick 33 Harry Potter characters. Then only a few excerpts from the Harry Potter books or fan fictions are manually collected and used, together with the characters, as prompts to GPT-3 to create 13k captions.
Results on relevant photos are shown in the third row of Figure~\ref{fig:stylistic_captioning}. The model uses the correct names and image content, while sometimes making up plausible events that could give additional context to the image as if it was a scene in a book or a movie.
\\
\boldheader{Reviews}
We train a model to write captions like a customer writing a review. For training data, we gather publicly-available Amazon product reviews\footnote{\url{https://www.kaggle.com/datasets/bittlingmayer/amazonreviews}} and select positive reviews that are a  maximum of 40 tokens long.  
As shown in Figure~\ref{fig:stylistic_captioning} bottom row, the captions use a variety of language to write positive reviews of the items in the photos.

\section{Related Work}
\boldheader{Using Contrastive Models}
Many vision and language contrastive models have been constructed, including CLIP~\cite{clip}, ALIGN~\cite{align}, UniCL~\cite{yang2022unified} and OpenCLIP~\cite{openclip}, and recent multi-modal models that contain a contrastive training component~\cite{yu2022coca,florence,albef}. 
Typically these models are used either zero-shot, which is effective for image classification but challenging for more complex tasks like captioning or visual entailment~\cite{song2022clip,tewel2021zero,zeng2022socraticmodels}, or as feature extractors for down-stream tasks~\cite{shen2021much,khandelwal2022simple,gu2021open,fang2021clip2video,luo2022clip4clip,portillo2021straightforward,zhai2022lit,wortsman2022robust}. 
Our work offers a compromise between those two approaches by allowing models to be trained with only textual data, which substantially improves upon zero-shot performance without requiring annotated images. 

\boldheader{Zero-Shot Vision Using Language Models}
Several recent works have combined large language models with pre-trained vision models to perform vision tasks zero-shot. Methods include using reinforcement learning to learn how to generate text that matches a CLIP Embedding~\cite{esper}, using CLIP to guide inference in the LLM~\cite{tewel2022zerocap}, or using a pre-trained model to generate text describing an image to pass into the language model~\cite{zeng2022socraticmodels}.
Compared to these methods our approach of leveraging text training has several advantages. Fine-tuning on text-only data enables our model to learn task-specific details and subtleties that are challenging for fully zero-shot methods, such as the style of captions to be generated. Our approach also works effectively with smaller language models (\Name{} only uses 220M trainable parameters) which significantly reduces the computational demand.

\boldheader{Cross-Modal Transfer Learning}
Transfer learning has typically focused on transferring skills from one modality to the same modality. 
CROMA is an exception and uses a modality-invariant feature space to achieve transfer similar to our work, however, it is limited to classification tasks and is few-shot rather than zero-shot~\cite{Liang2021CrossModalGL}.
Pre-trained language models have been shown to learn skills that can transfer to new modalities~\cite{Lu2021PretrainedTA}, however, this will be ineffective for task-specific skills such as a desired captioning style or learning the space of output labels.
Several multi-modal/multi-task models have learned many tasks in different modalities simultaneously~\cite{Lu2022UnifiedIOAU,Wang2022UnifyingAT,Liang2022HighMMTTM,Kaiser2017OneMT} and could thus potentially transfer skills between them, with HighMMT in particular showing positive results~\cite{Liang2022HighMMTTM}.
Our work studies the more challenging zero-shot setting (meaning no training data in the target modality is available), and therefore requires all the needed skills to be learned from a modality different than the one used in evaluation.

Recently, Song \etal~\cite{song2022clip} use a similar vector-substitution trick with CLIP to train visual entailment models, however they do no use noise or other methods that address the modality gap.
Yu \etal~\cite{esper} use reinforcement learning to train a model to generate text that CLIP ranks as being close to input images, and text data to learn captioning styles, although they do not directly train on text versions of the vision tasks.
Concurrently with our work, Nukrai \etal~\cite{nukrai2022text} and Wei \etal~\cite{lidecap} propose text-only approaches leveraging CLIP with either Gaussian noise similar to \Name{}, or using a projection of the text embeddings. 
Our work does additional analysis, covers more tasks including experiments using data generated by a language model, and achieves better captioning results.

\boldheader{Domain Invariant Representations}
Using domain-invariant features to achieve out-of-domain generalization has a long history in transfer learning. 
Work in this area has shown such features can be built from multi-domain training data~\cite{wang2022generalizing,gulrajani2020search}, small amounts of labelled data in the target domain~\cite{daume2009frustratingly,tzeng2014deep}, and unsupervised data~\cite{wilson2020survey,sun2019unsupervised}.
Methods include using adversarial learning to remove domain-dependent features~\cite{ganin2016domain,li2018deep,tzeng2017adversarial}, using maximum mean discrepancy to ensure features are distributed similarly across multiple domains~\cite{li2018domain,borgwardt2006integrating} and various data augmentation approaches to prevent models from learning domain-dependent features~\cite{zhou2021domain,zhou2020deep,volpi2018generalizing,shankar2018generalizing}.
The effectiveness of Gaussian noise in making models robust to domain shifts in these features has also been observed in image classification~\cite{li2021simple}.
While we also use domain-invariant features, the domain shift we study is more extreme than what is typically studied due to the change in modalities, and we show large-scale contrastive models can be an effective source of invariant features if used correctly.

\boldheader{Stylistic Captioning}
Stylistic captioning models can be built by authoring captions of the desired style~\cite{Mathews2016SentiCapGI,Gan2017StyleNetGA,Gurari2020CaptioningIT,Shuster2019EngagingIC} and applying standard captioning methods.
However, since creating such annotations is expensive, many stylistic captioning methods additionally transfer from captions with other styles by pre-training or multi-tasking~\cite{Mathews2016SentiCapGI,Nezami2018SentiAttendIC,You2018ImageCA}.
Other methods have combined unstylized captioning data with text data in the desired style through methods such as adversarial learning~\cite{Chen2017ShowAA}, multi-tasking with language modelling~\cite{Gan2017StyleNetGA}, or factoring caption writing into style and context components so that the style component can be learned from the text~\cite{Gan2017StyleNetGA,memcap}. 
Most similar to our work, Tan \etal~\cite{Tan2022DetachAA} train a model to generate text from either images or text using a shared encoding space and learned style embeddings. 
Unlike these methods, our approach does not require the use of any paired image/caption data. 

\section{Conclusion}
We have shown that the multi-modal semantic vector space learned by contrastive models can be used for cross-modal generalization through \Name{}, and studied its sensitivity and what improvements can be made with trained adapters. We have also conducted experiments on multiple vision and language tasks and demonstrated a specific application to stylistic captioning.
Beyond stylistic captioning, \Name{} is applicable to many other cases where training data is abundant in one modality but scarce in another. Possible use-cases include: training a captioning model for 3D scenes using image captioning data; training a model to summarize a video using text summarization data; and training a model to perform tasks like VQA or captioning for less-studied modalities like tables, graphs, or sensors without having to annotate additional data for all modalities.
As more powerful contrastive models that span more modalities are trained, we expect \Name{} to yield better results and gain more use cases.



{\small
\bibliographystyle{ieee_fullname}
\bibliography{egbib}
}

\clearpage

\setcounter{figure}{0}
 \setcounter{table}{0}
 \setcounter{footnote}{0}
 \setcounter{page}{1}
 \setcounter{section}{0}
 
\twocolumn[\section*{Appendix}
]


\section{Hyperparameters}
\label{sect:appendix-1}
\label{sect:hyperparameters}

For all tasks, we fine-tune our model with the Adam optimizer~\cite{kingma2014adam} with a linear decaying learning rate starting at $3\text{e-}4$, $\beta_1=0.9$ and $\beta_1=0.999$, batch size of 128, and train for 8 epochs. We use beam search with a beam size of 5 for evaluations. 
When tuning the noise level, we select 0.04 for VQA, 0.08 for visual entailment and visual news, 0.14 for captioning in the single caption setting, and 0.04 for captioning in the multiple captioning setting.

\section{Detailed Results}
\label{sect:appendix-2}

To facilitate more detailed comparisons with other works, we present results across more metrics of our evaluated datasets. In all tables, upper bounds that use images are shown above the dashed line.
\\

\begin{table}[]
    \tablefont
    \centering
    \begin{tabular}{l c c c c c}
        \toprule
         Model  & Mode & B-4 & M & C  & S \\
        \midrule
\ImageModel & - & 34.4 & 27.8 & 113.2 & 20.4 \\
\TunedModel{} & S & 28.6 & 25.2 & 95.4 & 18.1 \\
\TunedModel{} & M & 29.5 & 25.6 & 98.4 & 18.3 \\  
\hdashline
ESPER Style~\cite{esper} & - & 21.9 & 21.9 & 78.2 & - \\
\Baseline{} & S & 4.2 & 12.2 & 16.4 & 6.5 \\
\Baseline{} & M & 21.9 & 20.6 & 68.7 & 13.5 \\
\Name{} & S & 22.1 & 23.7 & 81.2 & 17.7 \\
\Name{} & M & 29.5 & 25.7 & 97.8 & 18.3 \\


    \bottomrule
    \end{tabular}
    \\
    \caption{Results on the caption test set in single-caption setting and multiple captioning setting, M indicates the multiple caption setting and S indicates the single caption setting.}. 
    \vspace{-0.2cm}
    \label{tab:captioning}
\end{table}

\begin{table}[]
    \tablefont
    \centering
    \begin{tabular}{l c c c c}
        \toprule
        Model & Yes/No & Num. & Other & All   \\
        \midrule
\ImageModel & 83.2 & 44.8 & 54.9 & 65.4 \\ 
\TunedModel & 79.4 & 43.4 & 51.1 & 61.9 \\ \hdashline 
TAP-C$_{ViT-B/16}$~\cite{song2022clip} & 71.4 & 20.9 & 18.6 & 38.7 \\
\NoiseModel & 77.1 & 42.1 & 48.6 & 59.6 \\
\Baseline & 78.6 & 40.6 & 49.0 & 60.2 \\
\bottomrule
    \end{tabular}
    \caption{Results on the \vqat{} test-dev set.}
    \label{tab:vqa}
\end{table}
\begin{table}[]
    \tablefont
    \centering
    \begin{tabular}{l c c c c}
        \toprule
        Model & Yes/No & Num. & Other & All   \\
        \midrule
\ImageModel & 80.4 & 48.4 & 64.1 & 67.9 \\ 
\TunedModel & 78.2 & 46.0 & 59.5 & 64.3 \\ \hdashline
\Name{} & 74.9 & 45.2 & 59.2 & 62.9 \\
\Baseline & 76.8 & 36.8 & 53.9 & 59.8 \\
\bottomrule
    \end{tabular}
    \caption{Results on the \vqae{} validation set.}
    \label{tab:evqa}
\end{table}
\begin{table}[]
    \tablefont
    \centering
    \begin{tabular}{l c c}
        \toprule
        Model & Val & Test \\
        \midrule
        \ImageModel & 77.0 & 77.7 \\ 
        \Name{} w/Tuned Noise & 75.9 & 75.9 \\ \hdashline
        CLIP Classifier~\cite{song2022clip} & 67.2 & 66.6 \\
        \Name{} & 75.9 & 75.9 \\ 
        \Baseline{} & 68.7 & 68.2 \\ 
    \bottomrule
    \end{tabular}
    \caption{Results on the visual entailment test and validation set.}
    \label{tab:visual_entailment}
\end{table}
\begin{table}[]
    \tablefont
    \centering
    \begin{tabular}{l c c c c}
        \toprule
        Model & B-4 & M & R & C \\
        \midrule
        VNC w/Images~\cite{visual_news} & 5.3 & 8.2 & 17.9 & 50.5 \\ 
        CLOSE w/Images & 9.3 & 10.9 & 25 & 105.7 \\
        \hdashline
        CLOSE & 5.4 & 8.2 & 19.7 & 80.8 \\
        CLOSE w/o Noise & 2.1 & 4.9 & 12.7 & 32.1 \\
    \bottomrule
    \end{tabular}
    \caption{Results on the visual news test set.}
    \label{tab:visual_news}
\end{table}

\boldheader{Captioning}
We present results in Table~\ref{tab:captioning} for BLEU-4~\cite{bleu}, METEOR~\cite{meteor}, \cider{}~\cite{cider} and \spice{}~\cite{spice}.

\boldheader{VQA}
We present results by question-type for \vqat{} in Table~\ref{tab:vqa} and \vqae{} in Table~\ref{tab:evqa}.

\boldheader{Visual Entailment}
We present visual entailment results on the test and dev set in Table~\ref{tab:visual_entailment}.

\boldheader{Visual News}
We present results with BLEU-4~\cite{bleu}, METEOR~\cite{meteor}, ROUGE~\cite{rouge} and \cider{}~\cite{cider} following~\cite{visual_news} in Table~\ref{tab:visual_news}. To the best of our knowledge the previous best reported results is from Liu \etal~\cite{visual_news} which does not make use of a pre-trained language model like \Name{} does. Qualitative results are show Section~\ref{sect:visual_news_examples}.

\section{Generating Synthetic Captions using Language Models}
\label{sect:appendix-3}

\begin{figure}
    \centering
    \includegraphics[width=0.5\textwidth]{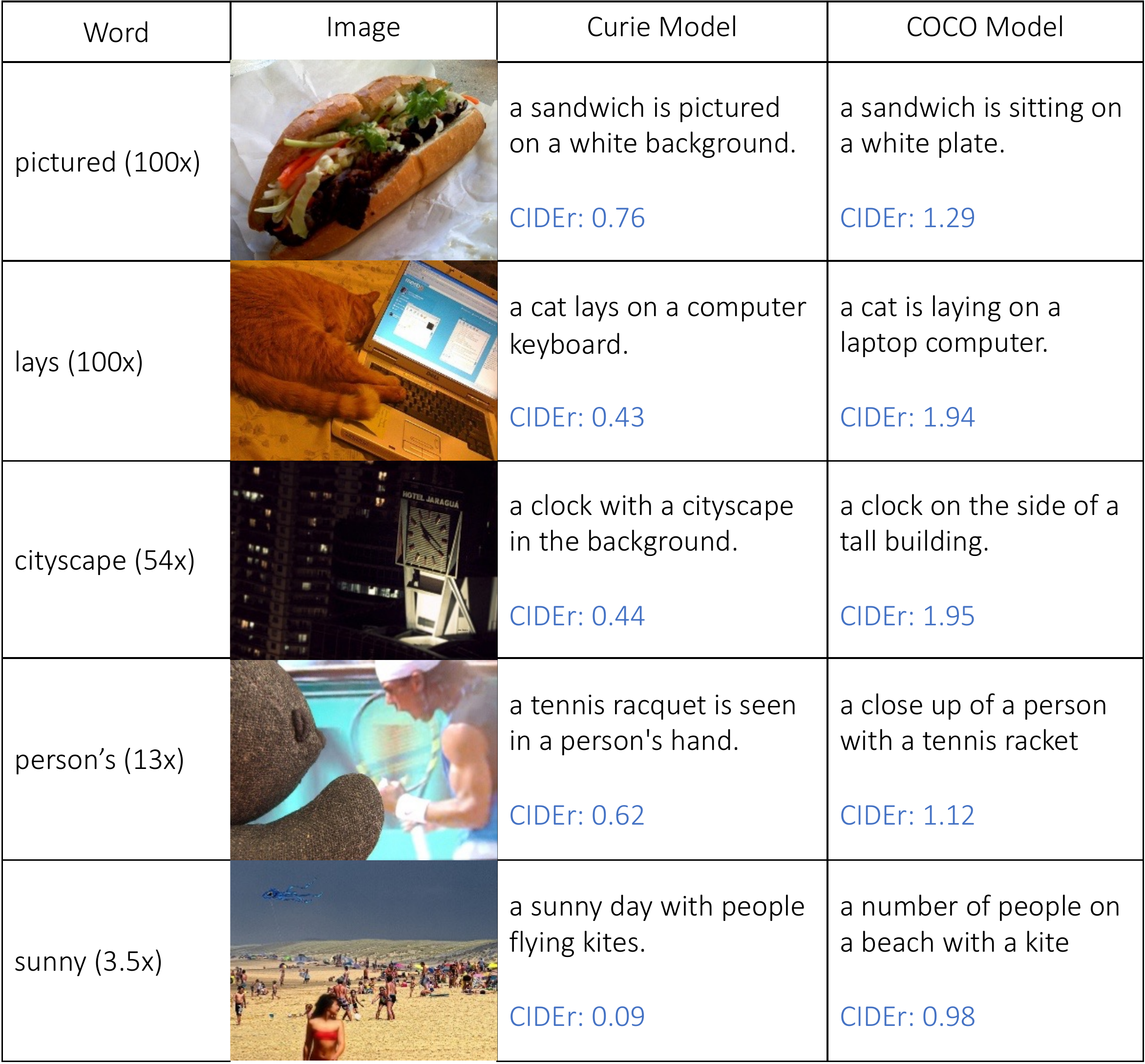}
    \caption{Examples of words that are over-produced by the captioning model trained on the \curie{} synthetic captions relative to the model trained on the \coco{} captions. The first column shows the word and how much more common it is across captions generated for images in the \coco{} validation set. The remaining columns provide an example image and a caption from both models with the \cider{} score computed using human-annotated captions.}
    \label{fig:syn_captioning_examples}
\end{figure}

\begin{table}[]
    \centering
    \begin{tabular}{l c c}
    \toprule
         Model & Individual & Any  \\ \midrule
         \curie{} & 58.8 & 85.0 \\
         \gptj{} & 42.7 & 81.9 \\
         \bottomrule
    \end{tabular}
    \caption{How often generated captions contain the target keywords when generating synthetic captions using different language models. The second column shows the success rate for individual generations, and the third column shows how often any caption in the 5 captions generated per a prompt contain both keywords.}
    \label{tab:keyword_stats}
\end{table}

In this section, we give more details about how we generate captions using language models and the results from Section~\ref{sect:training_with_language_model_data}.
When generating captions, we use nucleus sampling~\cite{holtzman2019curious} at $p=0.95$ and a temperate of 1, which we find generally improves results.
It is not uncommon for the caption to fail to contain both input keywords, so we sample 5 captions for each prompt and then select a caption containing the keywords if one exists, and select one randomly otherwise. 
The in-context example captions are prefixed by randomly chosen words that exist within that caption (excluding stop words), and we use randomly selected captions from \coco{} training captions as the examples.
During sampling, we randomly shuffle both the order of the in-context examples and what keywords are used as prefixes for those examples to improve the diversity of the outputs.
If doing unigram sampling, we keep track of the distribution of words found in the captions generated so far, and sample new keywords in proportion to how under-represented they are, while never sampling over-represented words.

Statistics for how often the input keywords are correctly included in the caption are shown in Table~\ref{tab:keyword_stats}. The success rate is less than 60\%, although selecting from 5 generations brings the success rate up considerably. \gptj{} is worse than \curie{}, but sampling extra captions helps make up for this deficiency.
Future work could integrate a constrained beam search method to address this difficulty~\cite{lu2021neurologic}.

We find that about 10\% of \gptj{} captions are not coherent or do not describe a visual scene, while these kinds of captions almost never occur with \curie{}.
Overall, for \gptj{}, producing 100k captions took about 50 GPU hours using a NVIDIA RTX A6000. For \curie{}, each generation requires approximately 500 tokens per a query, so the total cost was about 100\$\footnote{At the current rate of 0.002\$ per 1k tokens on 11/16/2022}. Both methods are far cheaper than annotating data.

As discussed, we observe stylistic differences occur between models trained on synthetic captions and models trained on \coco{} captions. A particular issue is that, while unigram sampling prevents words becoming under-represented, it still allows some words to become over-represented if the language model has a natural tendency to generate them. Figure~\ref{fig:syn_captioning_examples} contains some examples where the model trained on \curie{} captions uses words like ``pictured", ``lays"  or ``cityscape" that almost never occur in \coco{} captions and thus lead to low quantitative scores even when used correctly. Interestingly, we find \gptj{} is not as affected by this issue, which likely stems from differences in what data the language model was trained on.
Nevertheless, the captions do still correspond well to the image content, as shown by reasonably good captioning scores despite these stylistic issues, showing it is possible to learn captioning using only synthetic data.

\section{The Relationship Between Image and Text Vectors}
\label{sect:appendix-4}

\begin{figure*}[h]
    \centering
    \includegraphics[width=0.8\textwidth]{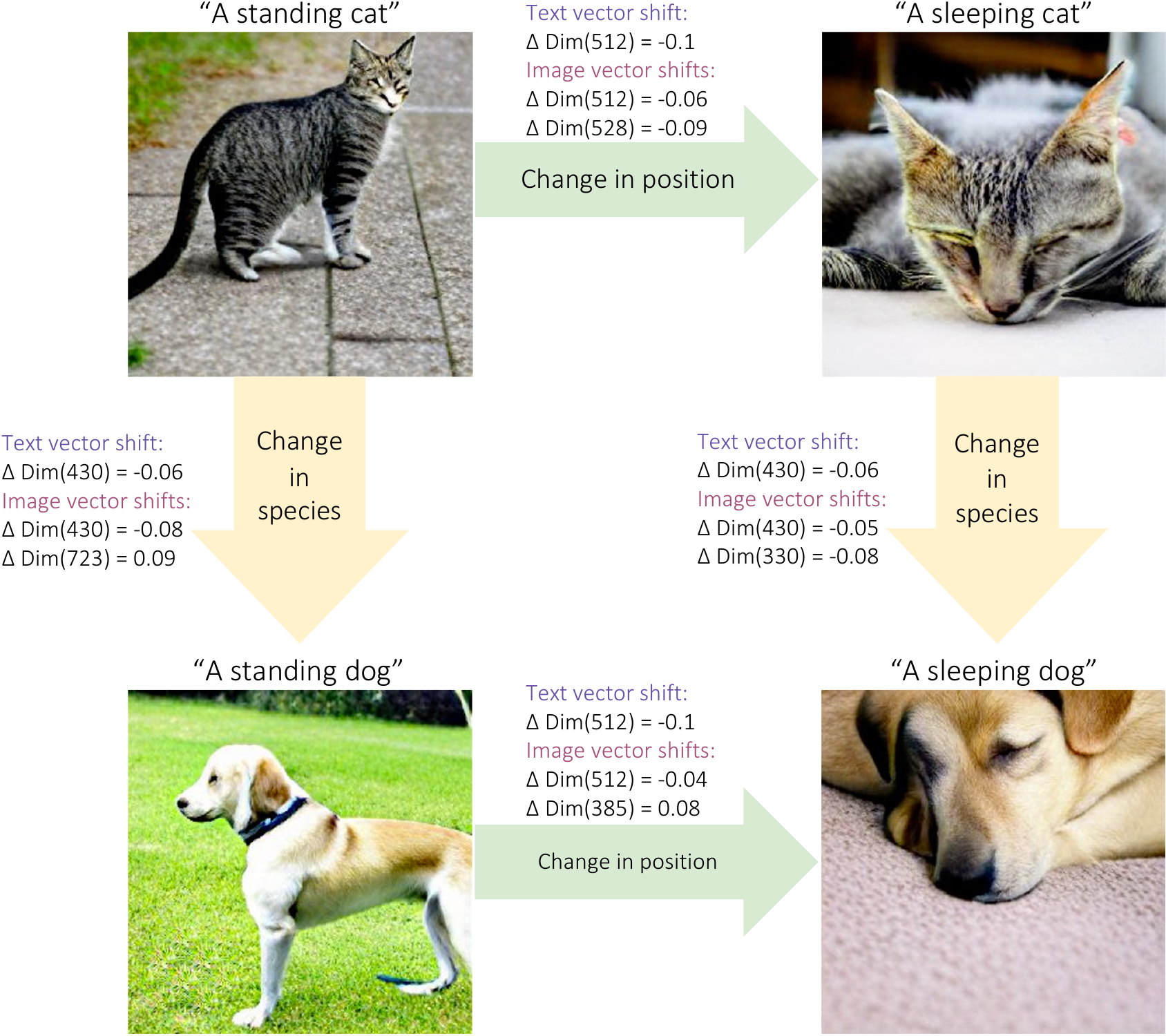}
    \caption{An example of how image/text feature vectors shift with a specific change in species (vertically) or position (horizontally). Text adjacent to each arrow shows any significant changes in the text (purple) or image (red) vector that occurred because of the shift.
    }
    \vspace{-0.2cm}
    \label{fig:vector_shifts}
\end{figure*}

\begin{figure*}
    \centering
    \begin{subfigure}[b]{0.32\textwidth}
         \centering
         \includegraphics[width=\textwidth]{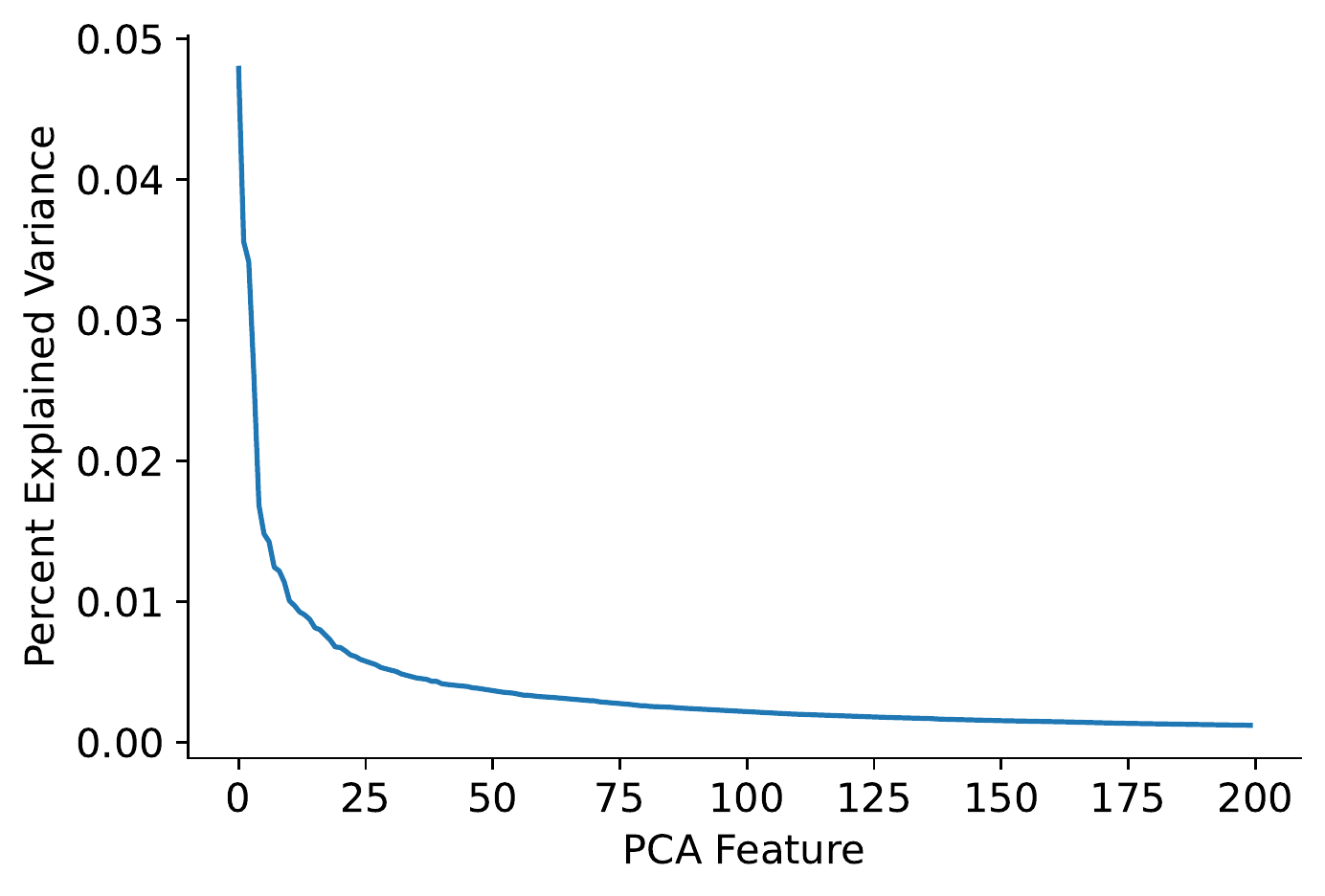}
     \end{subfigure}
    \begin{subfigure}[b]{0.32\textwidth}
         \centering
         \includegraphics[width=\textwidth]{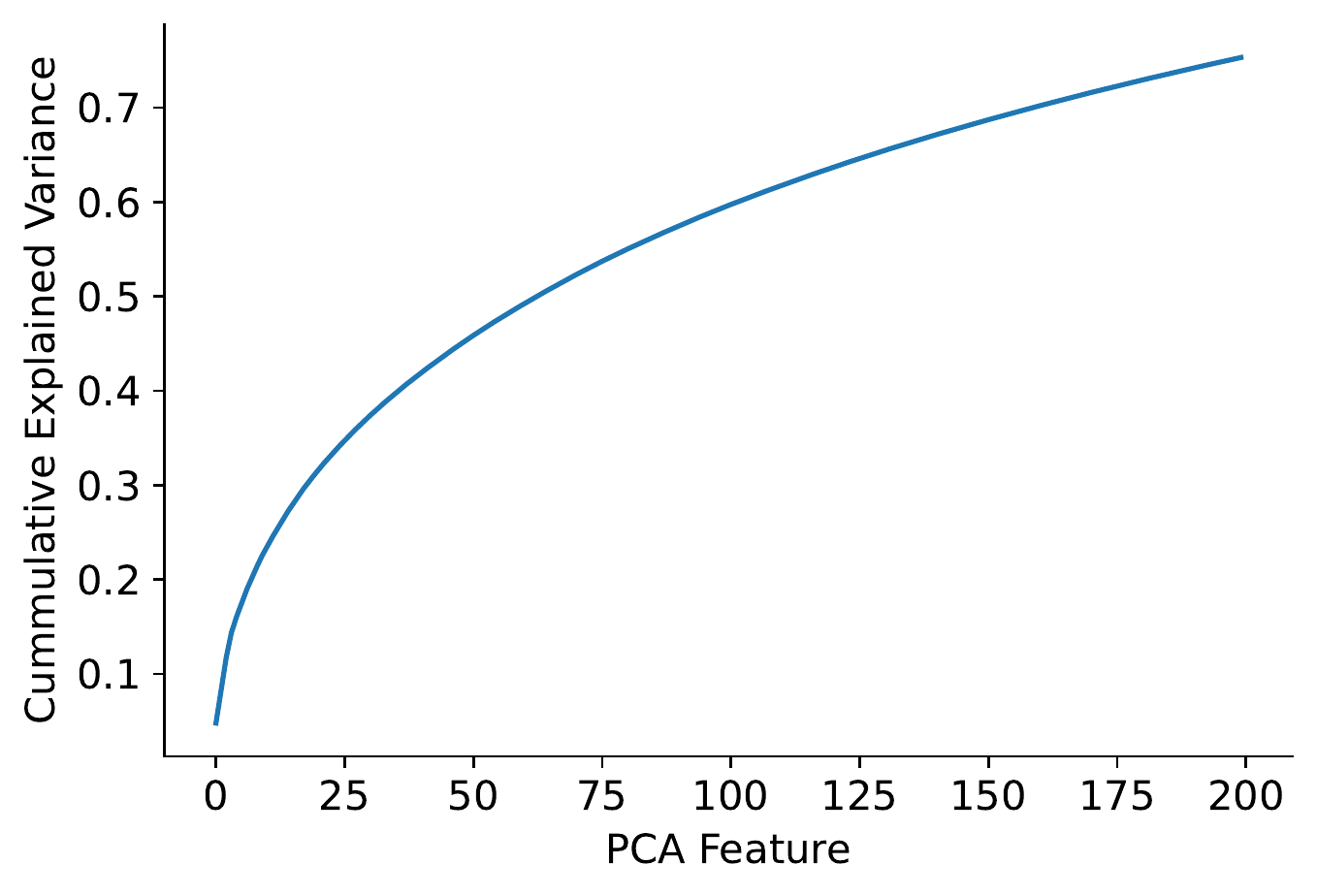}
     \end{subfigure}
    \begin{subfigure}[b]{0.32\textwidth}
         \centering
         \includegraphics[width=\textwidth]{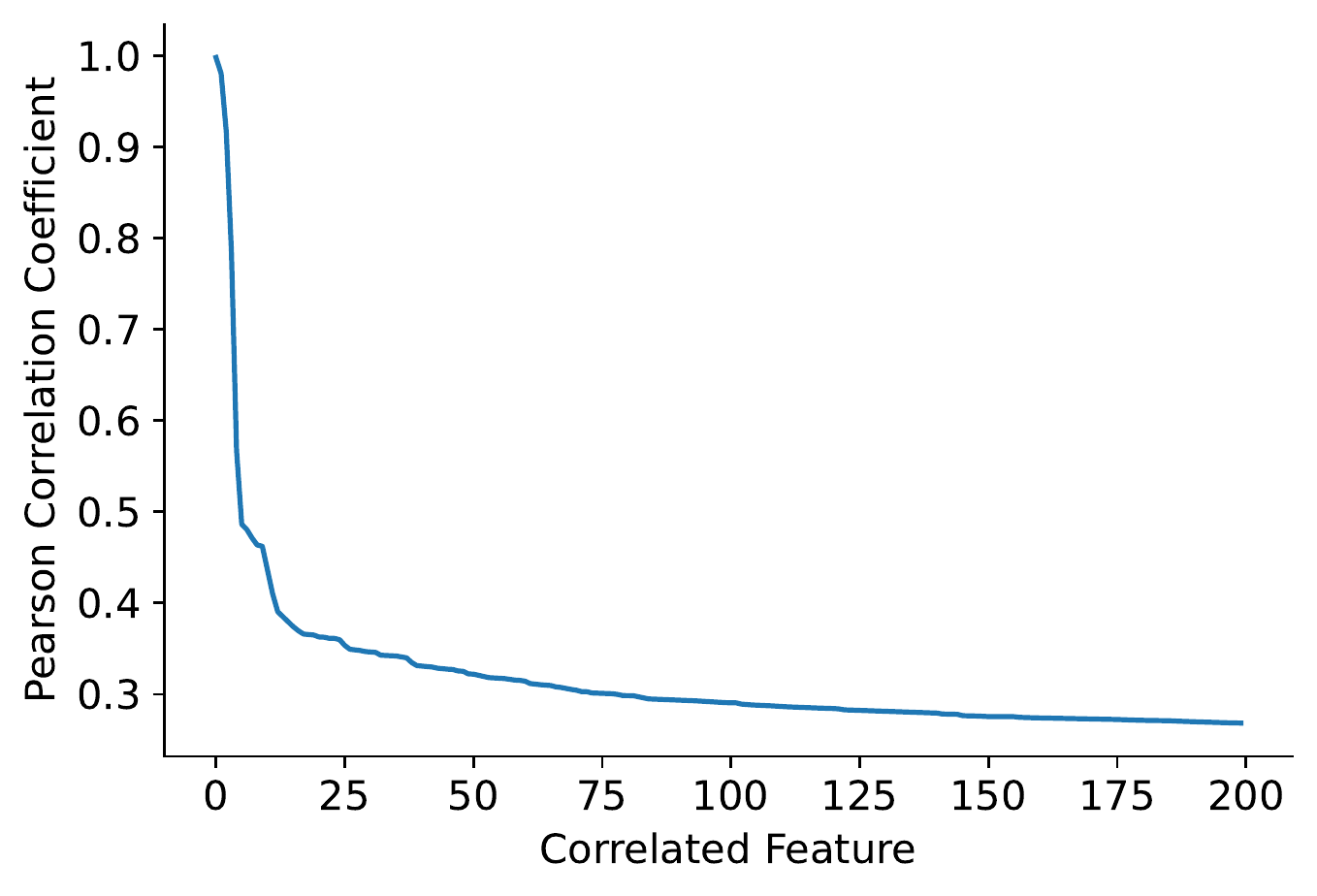}
     \end{subfigure}
    \caption{Plots analyzing the differences between image and text vectors for image/caption pairs in \coco{} captions. Only the first 200 features are shown.}
    \label{fig:vector_shifts_plots}
\end{figure*}

We perform a small case study by selecting four image/caption pairs that represent two different semantic changes in terms of animal species and positions (the result is shown in Figure~\ref{fig:vector_shifts}) and examine how the image or text vectors shift according to these changes. We observe that text vectors move more consistently when either the species or positions of the animals change. This disparity is likely due to random shifts in image semantics that correlate with conceptual changes in the text, such as subtle alterations in the animals' appearance, textures, or background.

We further analyze how image and text vectors typically differ by computing the differences between image/text pairs in an auxiliary corpus of COCO. We center these differences and apply PCA. The first two plots in Figure~\ref{fig:vector_shifts_plots} show that the first few PCA dimensions explain a large portion of the variance in these differences, showing that differences often occur in similar directions. We also plot the Pearson correlation coefficient for the most related features in the third plot, showing that a number of these features are highly correlated. Indeed, image/text pairs tend to move in a structured manner that follows a particular "shape". We capture this subtle relationship by studying the covariance matrix of the differences between text-image vectors. We then modify our Gaussian noise that is added to the text during training to better simulate this co-movement.

\section{Visual News Qualitative Examples}
\label{sect:appendix-5}

\label{sect:visual_news_examples}
We show some qualitative examples for visual news in Figure~\ref{fig:visual_news_fig}. We observe that close to $50\%$ of time, the predicted captions can be more descriptive (i.e., they can include more details), indicating there is room for this visual news captioner to grow. There are also some cases in which the predicted captions are better than the ones provided by human (the target captions). But overall, the general sense of both the news images and articles are present in the captions produced by CLOSE.

\begin{figure*}[p]
    \centering
    \subfloat{%
        \centering
        \includegraphics[width=0.98\textwidth]{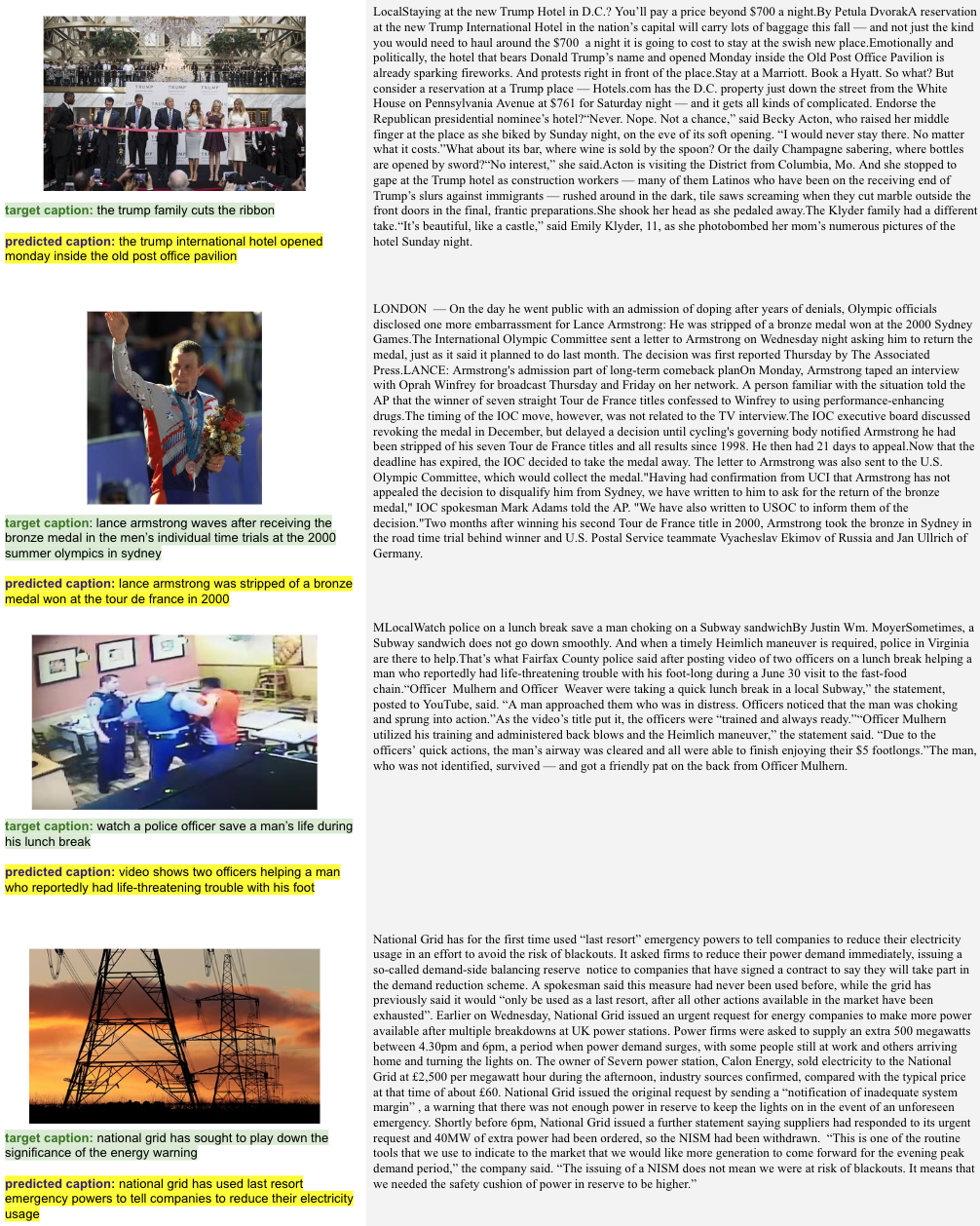}
    }
\end{figure*}
\begin{figure*}[p]
    \ContinuedFloat
    \centering
    \subfloat{%
        \centering
        \includegraphics[width=0.98\textwidth]{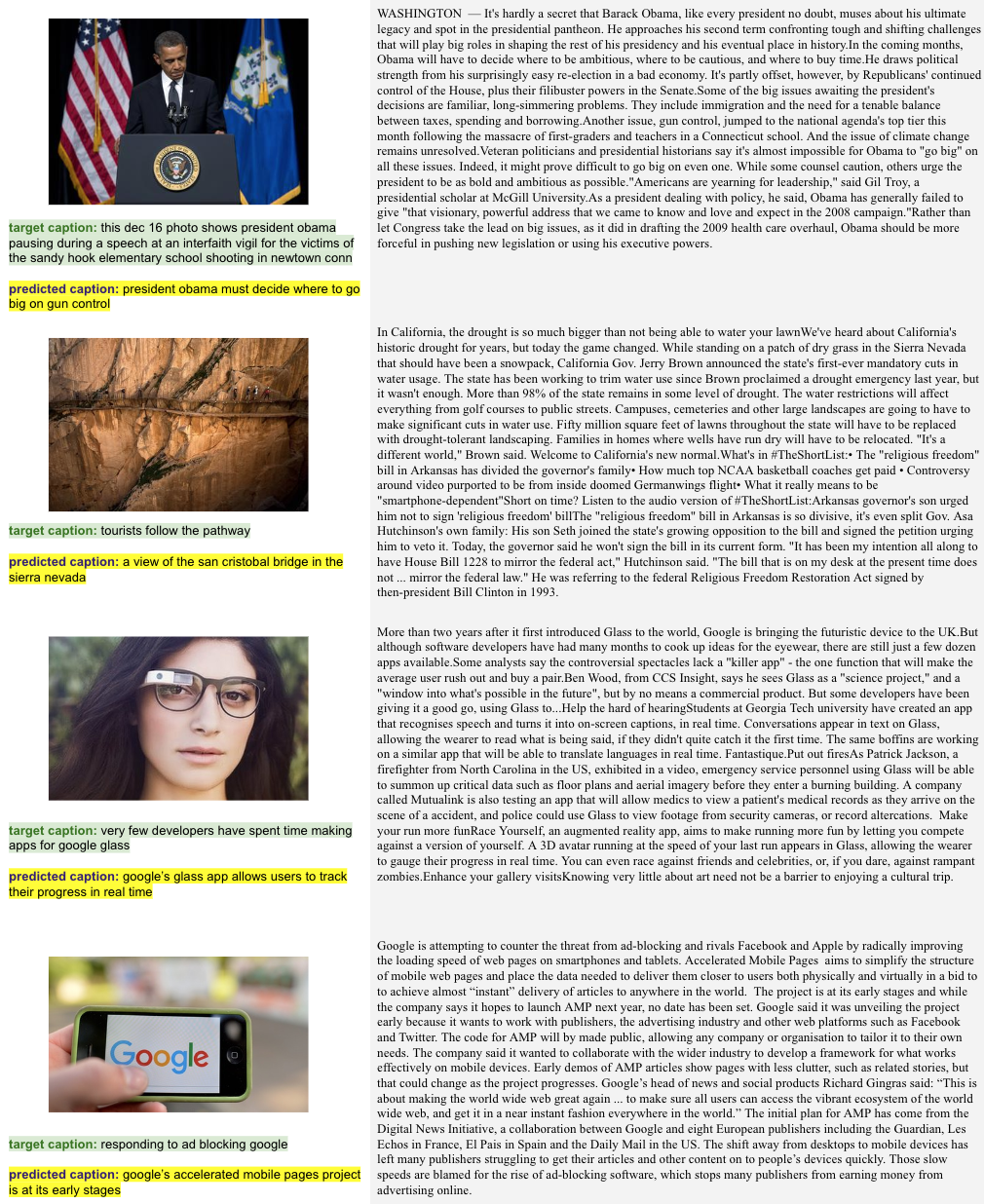}
    }
    \caption{Examples of visual news captions produced by CLOSE trained on text captions and news articles alone, and then applied zero-shot to news images and articles.}
    \label{fig:visual_news_fig}   
\end{figure*}

\end{document}